# Quantum-Inspired Evolutionary Algorithms for Feature Subset Selection: A Comprehensive Survey

**Yelleti Vivek[1,2], Vadlamani Ravi[1*], P. Radha Krishna[2]**

[1]Centre for Artificial Intelligence and Machine Learning

Institute for Development and Research in Banking Technology,

Castle Hills Road #1, Masab Tank, Hyderabad-500076, India

[2]Department of Computer Science and Engineering,

National Institute of Technology, Warangal-506004, India

yvivek@idrbt.ac.in; vravi@idrbt.ac.in; prkrishna@nitw.ac.in

## Abstract

The clever hybridization of quantum computing concepts and evolutionary algorithms (EAs) resulted in a new field called quantum-inspired evolutionary algorithms (QIEAs). Unlike traditional EAs, QIEAs employ quantum bits to adopt a probabilistic representation of the state of a feature in a given solution. This unprecedented feature enables them to achieve better diversity and perform global search, effectively yielding a tradeoff between exploration and exploitation. We conducted a comprehensive survey across various publishers and gathered 56 papers. We thoroughly analyzed these publications, focusing on the novelty elements and types of heuristics employed by the extant quantum-inspired evolutionary algorithms (QIEAs) proposed to solve the feature subset selection (FSS) problem. Importantly, we provided a detailed analysis of the different types of objective functions and popular quantum gates, i.e., rotation gates, employed throughout the literature. Additionally, we suggested several open research problems to attract the attention of the researchers.

## 1. Introduction

With Web 2.0 and big data advancements, real-world problems have become more complex. This complexity further increased with a vast number of dimensions or features, which started appearing in diverse domains. In such cases, not all features may be equally significant, with some being irrelevant or redundant. Including such features diminishes the performance of the underlying model. Feature subset selection (FSS) [1,2] enables the mitigation of these pertinent issues by selecting highly discriminative, informative, and relevant feature subsets. FSS is a paramount step in the CRoss Industry Standard Process for Data Mining (CRISP-DM) [3] framework for building models for data mining, such as classification, regression, and clustering.

FSS can be accomplished through *filter*, *wrapper*, and *embedded* approaches. Among these approaches, wrapper approaches that employ evolutionary algorithms (EAs) are often preferred owing to their high accuracy and ability to account for feature interaction effects. The astonishing benefits of FSS are as follows: it enhances the comprehensibility of the models, reduces the model complexity, improves the training time, avoids overfitting, and often improves the model's performance. Accordingly, the resulting model becomes parsimonious. The omnipresent existence of big datasets in every domain made FSS a mandatory pre-processing step.

* corresponding author ; vravi@idrbt.ac.in

FSS is a combinatorial problem because, for a given set of *n* features, there are $2^n - 1$ total possible number of feature subsets. This means the search space consists of all these possible feature subsets. The primary goal is to find an efficient feature subset, i.e., yielding as much as the highest possible accuracy by selecting a very small number of features with no or minimal redundant features and information loss. Identifying the best feature subset requires evaluating all the possible feature subsets. But, this exhaustive search makes it as a brute force method becomes impractical as the number of features *n* becomes large, especially in big datasets. Metaheuristics, including EAs, have proven superior to traditional optimization methods in solving multifarious combinatorial and continuous optimization problems.

In recent years, EAs have garnered significant attention from diverse research communities. The primary strength of EAs lies in their ability to solve NP-hard problems effectively. Due to the wide range of applications and the imperative for stability and accuracy, numerous enhancements and hybridizations have been proposed to the classical metaheuristics subsuming EAs. The field has witnessed algorithms and their modifications inspired by evolutionary processes, nature, biology, physics, etc. This phenomenal growth is attributed to the famous No Free Lunch (NFL) Theorem [81], since there is no single algorithm to address issues pertinent to various disciplines. Enhancements to EAs for a specific set of problems may come at the expense of performance in solving other problems. Hence, intensive study and subsequent introduction of new modifications became essential to address a broad spectrum of real-world challenges. One such important variant of EAs is the quantum-inspired evolutionary algorithms (QIEAs) paradigm, which is borne out of the synergy between quantum computing concepts and the EAs.

It is important to note that there are two important paradigms of employing Quantum principles with Evolutionary algorithms, namely, (i) Quantum inspired Evolutionary algorithms (QIEAs) and (ii) Quantum Evolutionary Algorithms (QEAs). QIEAs are the set of algorithms where the quantum principles are mimicked in classical computers. However, QEAs are the set of algorithms that were run on the actual quantum computers. In this survey, we restricted our analysis to QIEAs only.

**Table 1: Differences between various feature selection methods**

| **Variable** | **Filter** | **Wrapper** | **Embedded** |
|---|---|---|---|
| **Feature interactions [75]** | Not well captured | Well captured | Captured better than filter |
| **Addressing redundancy [75-77]** | Can select non redundant subsets to some extent | Most likely to select non redundant subsets | Likely to select non redundant subsets |
| **Time Complexity [76]** | Very less | Very High | High |
| **Scalable [77]** | Yes | Yes | Yes |
| **Examples** | Chi-square, t-statistic, mutual information etc. | Metaheuristics Plus classifiers/regression models such as LR, KNN, SVM, etc. | LASSO, DT, etc. |

Enhancements in the computational capabilities of classical computers fall short in achieving the day-to-day life demands since Moore's law is expected to falter by 2025 owing to the limitations imposed by the atom's size [78-80]. Quantum computing, with its property of quantum parallelism, offers a breakthrough by enabling quantum bits (Qubits) to manage multiple processes concurrently. The novelty and potential of quantum computing provide new avenues for solving optimization problems by enhancing the extant EAs. QIEAs gained traction owing to their efficiency and better search capabilities, utilizing the ability of quantum parallelism to manage complex computations.

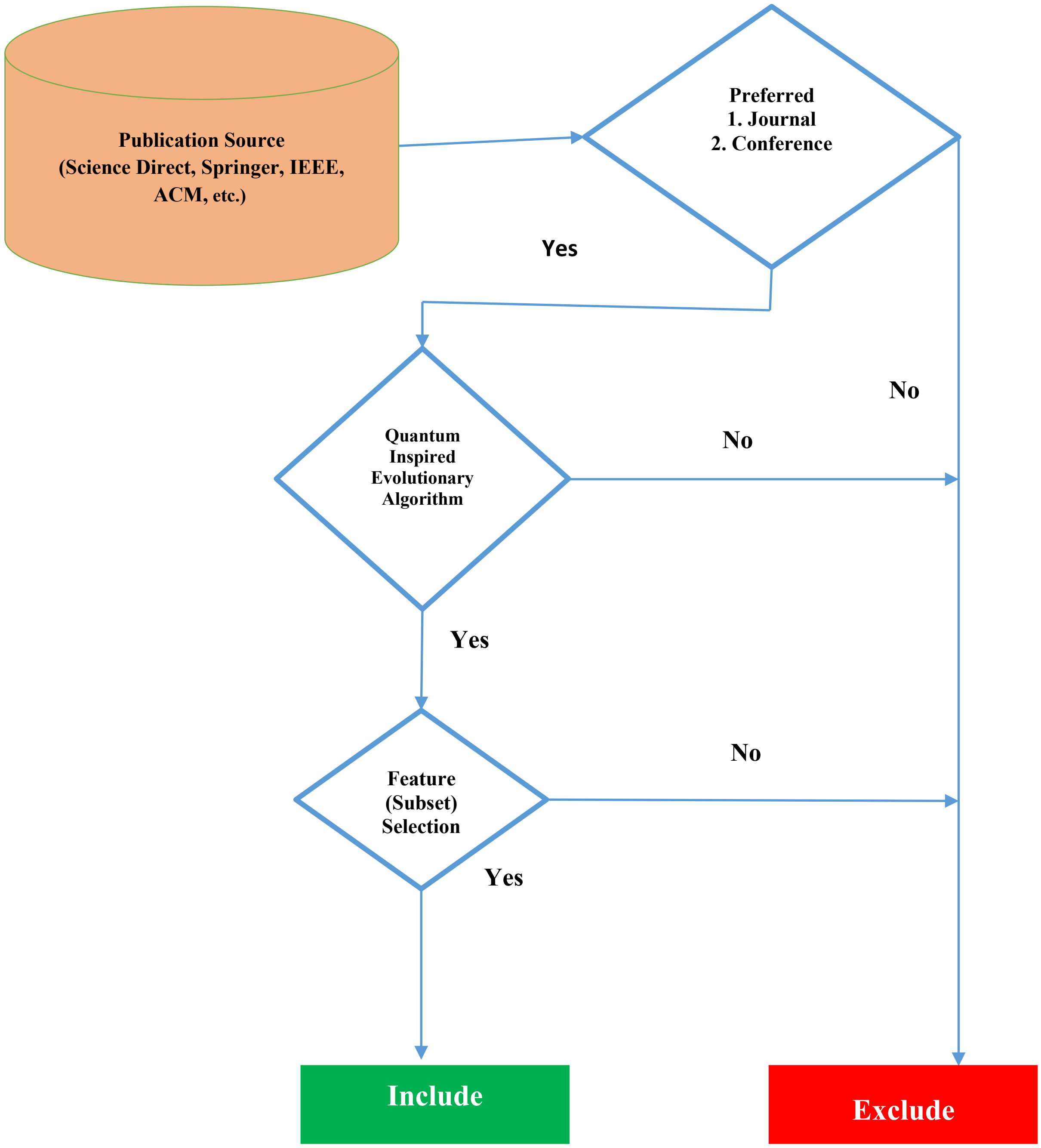

**Fig. 1. Methodology followed while choosing the articles for review**

### 1.1 Review Methodology

In the current work, we review and analyze the recent quantum-inspired metaheuristic algorithms and their applications developed between 2002 and 2024. We employed the following methodology, depicted in Fig. 1, while reviewing the articles. Our objective is to assist researchers in identifying current research trends, state-of-the-art technologies, and applications using this approach while also highlighting limitations and potential gaps in this field. The review includes works found in Web of Science, ScienceDirect, IEEE Xplore, MDPI, Emerald, Springer, and Google Scholar using keywords such as "quantum-inspired metaheuristic," "quantum-inspired," "quantum-inspired" + "attribute selection," "quantum" + "feature selection" "quantum-inspired" + "feature selection," and "quantum-inspired" + "evolutionary algorithm." Exclusion criteria included studies published in unknown journals and those with redundant information, prioritizing comprehensive studies over duplicative ones.

The major contributions of this survey are as follows:

- Conducted a comprehensive survey of 56 papers in the literature focusing on feature subset selection.
- Proposed taxonomy of quantum-based metaheuristics in the context of FSS.
- Identified the most important objective functions employed in the literature.
- Discussed 12 different types of rotation gates employed in the literature.
- Provided open research problems along interesting lines for the research community.

The remainder of the study is organized as follows: Section 2 discusses the motivation for the proposed survey. Section 3 covers the preliminaries of quantum computing, while popular gates are presented in Section 4. Section 5 introduces the proposed taxonomy of the QIEA FSS methods. The review of the works is presented in Section 6. In Section 7, we present the different extant types of objective functions and rotation gates. Section 8 outlines the open research problem, and the conclusion is presented in Section 9.

## 2. Motivation

We came across several survey papers [4-12] applying QIEAs to various domains (refer to Table 2). Nevertheless, to the best of our knowledge, no review work has been reported, concentrating particularly on FSS. Here, we presented a thorough analysis of the extant surveys to showcase the uniqueness and importance of the proposed survey as follows:

- Manju and Nigam [4] presented an exhaustive survey of quantum computational intelligence solving various business problems. They discussed motivation, challenges, and drawbacks involved in the extant approaches. Further, they presented open-research problems.
- Xiong et al. [5] discussed a comprehensive overview of quantum rotation gates and various types. Their review presented 21 different ways of implementing quantum rotation gates and the existing schemes that are involved in solving various research problems. Further, they also presented a comparative analysis among these variants.
- Savchuk and Fesenko [6] presented a comprehensive analysis describing the quantum computing model and ways of designing quantum efficient algorithms. Further, they presented the most recent articles and the challenges involved in solving various problems.
- Ross [7] reviewed quantum-inspired metaheuristics in the context of quantum computers solving various combinatorial and continuous optimization problems. They presented various challenges of implementing quantum-inspired metaheuristics in quantum computers.
- Uprety et al. [8] proposed a survey of the algorithms based on the quantum theory specific to information retrieval.
- Doung et al. [9] presented a comprehensive report on quantum-inspired machine learning for 6G applications.
- Araujo et al. [10] presented a review of quantum-inspired optimization methods for system reliability problems.

- Gharehchopogh [11] reviewed one research article where they comprehensively studied various quantum-inspired metaheuristics that solve various engineering problems and present future directions.
- Recently, Liu et al. [12] reviewed quantum-cognitive inspired algorithms solving sentiment analysis in various domains.

This work aims to review proposed QIEAs in the context of FSS over the last two decades. The objective is to help budding researchers and seasoned practitioners identify current research trends, cutting-edge technologies, and applications employing this approach while also highlighting their limitations and potential open research problems.

**Table 2: Summary of the various surveys related to Quantum-inspired algorithms**

| **Authors (year)** | **Theme of the survey** | **FSS oriented papers included with (count)** |
|---|---|---|
| Manju and Nigam [4] (2014) | Quantum-inspired computational intelligence | ✓ (3) |
| Xiong et al. [5] (2018) | Comprehensive survey on quantum rotation gate | ✓ (1) |
| Savchuk and Fesenko [6] (2019) | Quantum computing | × |
| Ross [7] (2019) | Quantum-inspired metaheuristics in real quantum computers | ✓ (1) |
| Uprety et al. [8] (2020) | Quantum theory based approaches for information retrieval | × |
| Doung et al. [9] (2022) | Quantum inspired Machine learning for 6G applications | × |
| Araujo et al. [10] (2022) | Quantum optimization algorithms for reliability problems | × |
| Gharehchopogh [11] (2023) | Quantum-inspired metaheuristics comprehensive survey | ✓ (7) |
| Liu et al. [12] (2023) | Quantum cognitively inspired algorithms for Sentiment Analysis | × |
| Current study | Quantum-inspired evolutionary algorithms for feature selection | ✓ (56) |

# 3. Preliminaries of Quantum Computing

By utilizing the ideas of quantum computing to enhance the performance of traditional metaheuristics, Han and Kim [13] introduced a novel paradigm called QIEA in their seminal work. Metaheuristic algorithms have a long history of maintaining an optimal balance between exploration (global search) and exploitation (local search). Since giving priority to one over the other often comes at the expense of the other, it is important to find a balance between them. The incorporation of quantum computing principles into these algorithms seeks to improve global search capabilities without compromising the effectiveness of the exploitation phase.

In contrast to classical computing, where basic units of information are simple bits limited to values of either 0 or 1, quantum computing utilizes *Qubits* as its fundamental unit. Similar to classical bits, a Qubit can exist in either a 0 or 1 state. However, unlike classical bits, Qubits can exist in a superposition of both states simultaneously. This unique property of Qubits enables them to execute numerous processes more rapidly. Mathematically, a Qubit can be represented as follows:

$$Q = \begin{bmatrix} \alpha_1 & \alpha_2 & \dots & \alpha_n \\ \beta_1 & \beta_2 & \dots & \beta_n \end{bmatrix} \quad (1)$$

An N-Qubit register can simultaneously hold $2^N$ states, whereas an N-bit register in a conventional computer can only represent a single value out of $2^N$ possible values. This characteristic is fundamental to quantum parallelism, enabling an exponential improvement in efficiency.

The Qubit states are updated through employing various quantum gates including the NOT gate, CNOT gate, Rotation Gate, Hadamard Gate, etc. Notably, a quantum gate is a linear transformation and is by default it is reversible. It is represented by a unitary matrix, denoted as U. A complex square matrix U is considered unitary if its adjoint U† and its inverse $U^{-1}$ are identical, and both the rows and columns of U are orthonormal.

Updating the values of $\alpha_i$ and $\beta_i$ are the cartesian coordinates / probabilities using the Rotation gate can be observed in Figure 1 . It is interesting to note that the three-dimensional model of Qubits in the Bloch sphere (see Fig. 2) is often reduced to two dimensions in the quantum-inspired metaheuristics literature. This simplified representation focuses on the rotation angle and probability amplitudes before and after rotation. Furthermore, classical computers simulate the measurement of quantum bits while accounting for probability amplitudes.

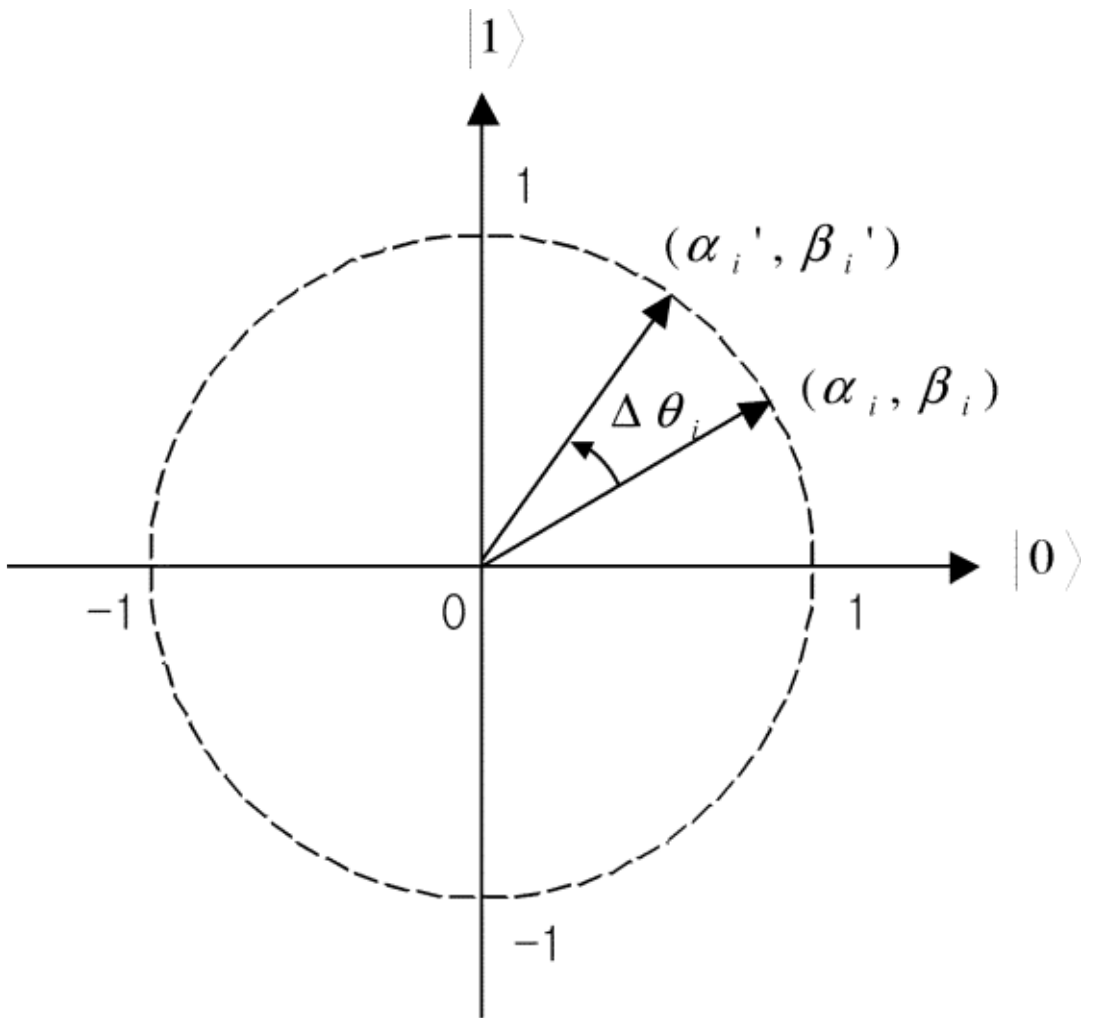

**Fig. 2. Bloch sphere or polar point representation of Qubits [13]**

## 4. Popular Quantum Gates employed in QIEA

Any operations on a quantum computer can be carried out using a set of universal quantum gates. This implies that a finite sequence of gates from this set can be used to represent any other unitary operation. Since the number of possible quantum gates is uncountable and the number of finite sequences from a finite collection is countable, it is technically impossible to accomplish this with a countable set of gates. We simply require that each quantum operation be able to effectively approximate by a series of quantum gates from the finite set to accomplish this task. Additionally, the Solovay–Kitaev theorem [19] guarantees the efficiency of this approximation for unitaries on a constant number of Qubits. Through this theorem, it is possible to ascertain whether a set of quantum gates is universal can be accomplished using group theory methods or by examining its relation to (approximate) unitary t-designs.

Operators or gates are used to implement mathematical and logical operations on Qubits that are represented as vectors. A matrix representation for an operator is a simple way to understand how a quantum system can be transformed from one state to another.

### 4.1 Rotation Gate

A Quantum rotation gate is the transformation of a single Qubit rotation based on a rotation angle. It is the most often used gate in the literature and is presented as,

$$\mathrm{R}(\theta) = \begin{bmatrix} \cos(\theta) & -\sin(\theta) \\ \sin(\theta) & \cos(\theta) \end{bmatrix} \tag{2}$$

Where ϴ is the rotation gate chosen in various ways such as either based on the reference table or adjusted dynamically. More details regarding this are provided in the Section 7.2.

### 4.2 Hadamard Gate

It is a single-bit Qubit operator and is presented in Eq. 3. It transforms basic states of \0 and \1 into superposition states.

$$\mathrm{H} = \begin{bmatrix} \frac{1}{\sqrt{2}} & \frac{1}{\sqrt{2}} \\ \frac{1}{\sqrt{2}} & \frac{-1}{\sqrt{2}} \end{bmatrix} \tag{3}$$

### 4.3 Pauli X gate

Pauli X gate represents the rotation around the x-axis of the Bloch sphere by $\phi$ radians. This gate is the quantum equivalent of the NOT gate for classical computers.

$$\mathrm{X} = \begin{bmatrix} 0 & 1 \\ 1 & 0 \end{bmatrix} \tag{4}$$

### 4.4 CNOT gate

Controlled gates act on 2 or more Qubits, where one or more Qubits act as a control for some operation. For example, the controlled NOT gate (or CNOT or CX) acts on 2 Qubits and performs the NOT operation on the second Qubit only when the first Qubit is  presented in Eq. 5.

$$CNOT = \begin{bmatrix} 1 & 0 & 0 & 0 \\ 0 & 1 & 0 & 0 \\ 0 & 0 & 0 & 1 \\ 0 & 0 & 1 & 0 \end{bmatrix} \tag{5}$$

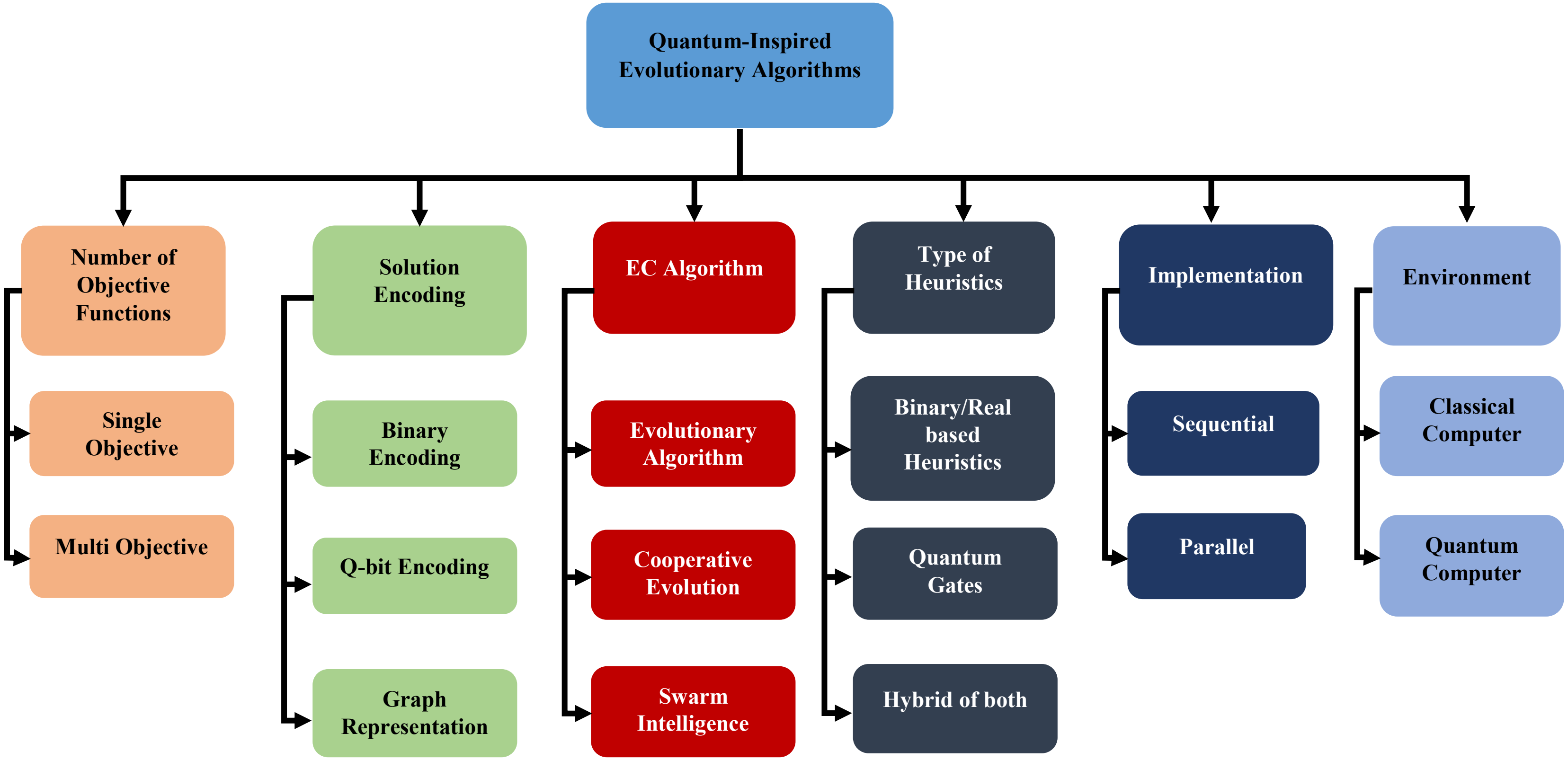

**Fig. 3. Taxonomy of the QIEA FSS algorithms**

## 5. Quantum Inspired Evolutionary Computing based FSS Taxonomy

This survey provides a comprehensive taxonomy (see Fig. 3) of the existing methods in various categories such as solution encoding, number of objective functions, type of algorithm, type of heuristic, and implementation. Based on the number of objective functions considered, there are two distinct categories: (i) Single objective and (ii) Multi-objective environment. Based on the solution encoding, there are three different categories: (i) Binary Encoding, (ii) Q-bit encoding, and (iii) Graph representation encoding. Based on the type of EC algorithm, there are three distinct techniques: (i) evolutionary algorithm, (ii) cooperative evolution, and (iii) swarm intelligence. When the underlying heuristic is considered, these can be (i) binary encoding or real encoding, (ii) quantum gates, and (iii) a combination of both. Based on the implementation, it could be either (i) a Parallel algorithm or (ii) a Sequential algorithm. Even though none of the methods in the literature were noticed to be parallel, we introduced this into the taxonomy owing to its importance. Based on the type of environment, optimization algorithms can be implemented in Quantum computers and Classical computers. Note that all of the algorithms that are reviewed in this survey are developed using classical computers. The year-wise trend analysis of the QIEAs is depicted in Fig. 4, and the publication house-wise analysis that published various QIEAs is depicted in Fig. 5.

## 6. REVIEW OF THE QIEA-BASED FSS WORKS

In this section, we presented a review of the extant approaches algorithm-wise (refer to Table 4) and the corresponding count is presented in Table 3. Further, we presented a detailed analysis by considering a few aspects such as (i) the objective of the study; (ii) whether the algorithm is a single or multi-objective optimization algorithm; (iii) whether implementation is sequential / parallel in nature; (iv) employed classifier; (v) targeted application; (vi) adaptive or non-adaptive; (vii) type of data partitioning that was employed during experimental analysis (holdout / k-fold cross validation / leave one out CV); (viii) developed in either classical or quantum computer; (ix) encoding scheme employed for the proposal (binary/ real / Qubit chromosome/graph representation) (x) gates employed in the respective study. This indeed helps us summarize the research directions taken by the researchers so far.

**Table 3: Algorithm-wise publication**

| Sl.No | Algorithm | Number of Publications |
|---|---|---|
| 1 | Genetic Algorithm (GA) | 20 |
| 2 | Particle Swarm Optimization (PSO) | 13 |
| 3 | Gravitational search algorithm | 3 |
| 4 | Differential Evolution (DE) | 2 |
| 5 | Whale optimization algorithm (WAO) | 2 |
| 6 | Avian navigation algorithm | 2 |
| 7 | Grey wolf algorithm | 2 |
| 8 | Grasshopper optimization algorithm | 1 |
| 9 | Immune clonal algorithm | 1 |
| 10 | Moth flame optimization algorithm | 1 |
| 11 | Hummingbird algorithm | 1 |
| 12 | Ant bee colony algorithm | 1 |
| 13 | Squirrel optimization algorithm | 1 |
| 14 | Owl optimization algorithm | 1 |
| 15 | Dwarf mongoose algorithm | 1 |
| 16 | Honey badger algorithm | 1 |
| 17 | Fruit fly algorithm | 1 |
| 18 | Immune clone algorithm | 1 |
| 19 | Shuffled frog leap algorithm | 1 |
| **Total** | | **56** |

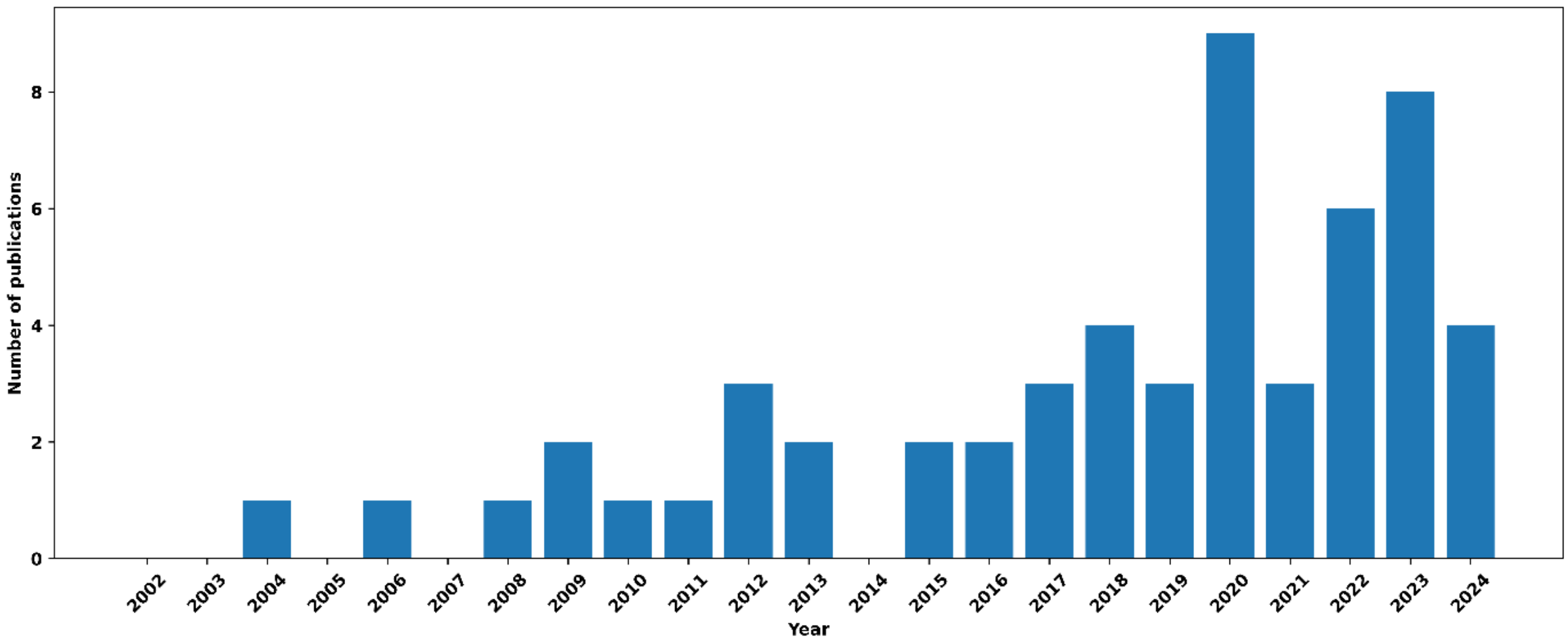

**Fig. 4 Year-wise trend of the publication in various avenues**

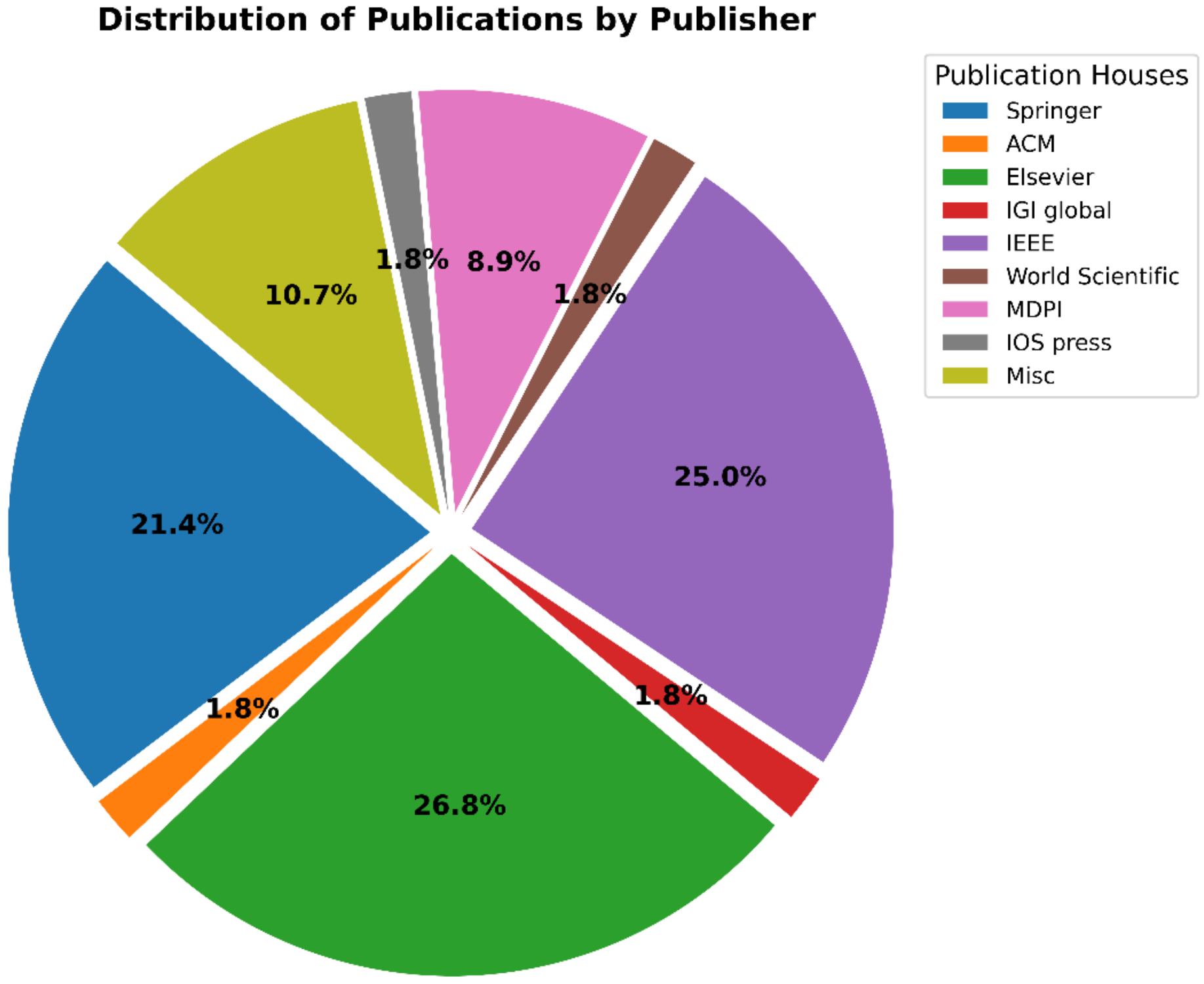

**Fig. 5 Publisher wise distribution of various QIEAs**

## 6.1 Variants of Quantum Genetic Algorithm

Zhang et al. [14] proposed a Quantum Genetic Algorithm (QGA) with a novel resemblance coefficient and applied it to radar-emitting signals datasets. Initially, the features are extracted from thwas generated radar emitting signals. Then, the resemblance coefficient was calculated, which subsumes the calculation of the correlation; the higher the value, the more it resembles those features. Importantly, thwas was calculated pair-wwase for all the features, and the highly correlated features are identified. In the next step, the class separability matrix was calculated to compute the separability between two classes and between two different signals, corresponding to computing the correlations of two dwastinct functions. The separability matrices are calculated for each algorithm, and now the task was reduced to identify the best feature subset and how the matrices are chosen. Now, the QGA was employed to determine the better combination. It embeds the implementation of a quantum rotation gate based on the lookup table.

Zhang et al. [15] proposed a Novel QGA (NQGA), where a novel strategy was proposed to update the rotation angle of quantum logic gates. The rotation angle was based on the multiplicative factor of the user-defined constant and the search direction function, which was obtained from the lookup table to determine the search direction of convergence. Later, each Qubit was updated with a certain probability, which was adaptive and fixed for an individual Qubit. Thwas enhances the search capability and prevents the algorithm from getting entrapped in the local optimal solution. Further, they implement a population catastrophe operation that could make the algorithm progress throughout the evolution without stagnating after a certain number of generations. Thwas operation also enhances the capability of the algorithm to jump out of local minima.

Chen and Zou [16] proposed a Quantum-inspired clone GA (QCGA) to improve text categorization by addressing the information overload wassue to filter out irrelevant information and automatic text categorization. Initially, the features were extracted, and then QCGA was executed for a certain number of generations, where it underwent clone, crossover, and mutation operations. They proposed a novel operation and named it clone operation, where the new population was generated by the best solutions found thus far. These clone solutions undergo crossover and mutation operations and will replace the worst pwerent solutions.

Wei and Ye [17] proposed a two-stage QGA-based method that works in two stages: (i) expected cross entropy (ECE) was employed to filter out the unimportant features, and (ii) the quantum genetic algorithm (QGA), where the quantum revolving gate was employed which behaves like quantum rotation gate. The angle of rotation was based on the reference table. Their approach demonstrated efficacy in the text categorization datasets, where it reduced more than 70% of the features with improved accuracy of more than 80%.

Abderrahim et al. [18] proposed a hybrid algorithm called GQASVM, which combines the Genetic Quantum Algorithm with the Support Vector Machines classifier. Their work address wassues in the high dimensional datasets. They employed quantum mutation and crossover, analogous to the classical flip-bit mutation and one-point crossover. However, these operations were performed in an individual's entire Qubit state. Further, the quantum inference was achieved by employing a quantum rotation gate. The angle of the rotation was kept under control since the larger rotation angle leads to getting the algorithm struck at a local optimal solution. They demonstrated the efficacy of their algorithm in the microarray datasets dealing with various cancer types (leukemia, breast, colon, ovarian, prostate, and lung).

Manju et al. [19] proposed a hybrid approach called ClusterQGA. Initially, clustering was used to select a small set of non-redundant representative genes. Next, QGA was applied to determine a minimal set of relevant and non-redundant genes. To avoid divergence or premature convergence, they employed small values for the rotation gate. The algorithm's stagnation was handled using catastrophe operations, where a set of best solutions was retained and the remaining solutions were reinitialized. Thwas approach helps avoid the algorithm from becoming trapped in a local optimal solution. Further, they proposed a novel fitness function by multiplying both the cardinality and accuracy of the model. Their approach demonstrated superiority in both binary and multi-class cancer microarray datasets available publicly.

Zhang and Wang [20] proposed QGA to correct the recognition rate of rolling bearing failure. They employed a quantum rotation gate, where the angle of rotation was decided based on the fitness function comparwason between the best individual to the current and the corresponding Qubit state. Further, through the dynamic strategy, the angle was adjusted to improve the convergence capability of the algorithm. They demonstrated the efficacy of their approach, and QGA improved to 86%.

Aranian et al. [21] proposed QGA to improve the accuracy and handle the curse of dimensionality in optical character recognition (OCR) on Persian handwritten letters. They initially generated the population by giving equal amplitude preferences to both states of the Qubit. Later, the offspring were generated by employing the rotation gate on the pwerent solutions. The rotation angle and the sign of the amplitude were decided based on comparing the fitness scores of the current solution with the best solution, which was referred to in the lookup table. They employed a quantum rotation gate operator and relied on the lookup table to decide the rotation angle. They demonstrated that by reducing the number of features by 19-49%, the accuracy of the model was improved by 7-31%.

Ramos and Vellasco [22] proposed a quantum-inspired evolutionary algorithm (QIEA) and applied it to the electroencephalography (EEG) signal data. Initially, the features were extracted from the EEG signal dataset, and thereafter, QIEA was employed to select the optimal feature subset. QIEA embeds with the quantum rotation gate, which was non-adaptive, and the rotation angle was a user-defined parameter. However, through their experiments, they validated that updating the rotation angle with 0.002pi and 0.0025pi achieves superior results.

Tayarani et al. [23] proposed a PCA-based Quantum evolutionary algorithm (PCA-QEA) for selecting the optimal number of feature subsets and identifying the personality trait of an individual. They proposed a novel Observation operator (O-operator) to generate the offspring. Once the offspring were generated, they will be combined with the pwerent solutions, and the best solutions will be decided upon by employing principal component analyswas (PCA). Here, PCA finds the direction that accounts for a large variance. The smaller the variance, the better the components and the better the solutions tend to be stable. They considered 2,998 fillers uttered by 120 different speakers in various conversations. They demonstrated that after performing feature selection, the accuracy improved to 74-82% in the particular trait.

Ramos and Vellasco [24] proposed a chaotic-based QIEA and named it CQIEA. They employed different chaotic maps such as burgers map, he map, Ikeda map, logwastic map, quadratic map, and Tinkerbell map and performed a comparative analyswas. Note that these chaotic maps were employed to update the rotation angle. Among all the chaotic maps, the logwastic map and Ikeda-driven rotation gate achieved higher accuracy than the QIEA [22] and the non-quantum algorithms. They further noticed that the convergence was greatly improved after employing chaotic maps, and exploration and exploitation were well balanced.

Ram et al. [25] proposed a quantum genetic algorithm (QGA) where a quantum rotation gate was employed to update the quantum bits. They employed the T-score technique to extract features from the microarray datasets. Further, they also employed quantum crossover and quantum mutation operators and applied them to a couple of datasets. They demonstrated their proposed algorithm outperformed the non-quantum counterpart of GA.

Chakraborty et al. [26] proposed a hybrid quantum feature selection algorithm (HQFSA) by employing the concept of a correlation coefficient based graph-theoretic classical approach. Their approach initially verifies whether the dataset was suitable for dimensionality reduction by employing quantum Oracle with CNOT operation. Based on their suitability, the process was continued by invoking HQFSA, where it can efficiently estimate high correlation values by using quantum parallel amplitude estimation and amplitude amplification techniques. They showed a detailed comparwason of HQFSA with some state-of-the-art feature selection algorithms regarding accuracy and CPU computational time measurement on some popular UCI datasets.

Kirar and Agarwal [27] proposed a combination of spectral graph and QGA. They employed quantum crossover and mutation operators followed by quantum population update operation. The quantum population update employs a rotation gate, but the rotation angle was computed in a different fashion. However, the rotation angle was computed as per the reference table. Further, they also introduced catastrophe operation, which was employed if there was no sign of improvement in fitness after a certain number of successive generations. Thwas helps avoid the problem of trapping in local optima. Their approach led to a reduction in the number of electrodes, reducing the cost and computation time needed to build BCI.

Azzam et al. [28] proposed a quantum genetic algorithm (QGA), which was employed to find the optimal neural network architecture over a set of possible model architectures. Further, QGA was also used for feature selection to remove the non-significant input parameters that did not contribute significantly to the output. The number of parameters was provided by the Bayesian regularization. The results demonstrated that the newly introduced regularization parameter improved the model's performance with a 75% reduction in the mean relative error compwered with a normal radial-baswas network.

Ling and Hao [29] proposed a parallel adaptive quantum genetic algorithm (NMIFS MOP-AQGA) by employing both normalized mutual information and cooperative evolution of multiple operators. It was a two-stage method, where the first stage NMI was employed to filter out important features and give them to the MOP-AQGA algorithm. The algorithm generates the offspring through the rotation gate, where the rotation angle was adjusted adaptively based on the algorithm's progression. They demonstrated the efficacy of the NMIFS MOP-AQGA on the intrusion detection datasets. Their approach has higher detection accuracy, lower false negative rate, and higher adaptive performance. However, they did not utilize any parallel / dwastributed frameworks in their setup. Hence, they were not truly parallel. They were just mimicking the behaviour.

Li et al. [30] proposed a quantum approximate optimization (QAO) algorithm for FSS by employing a graph theoretic approach. They created three unique graphs from the raw feature set based on three different metrics, namely, symmetric uncertainty, normalized mutual information, and correlation coefficient. The corresponding feature subset was generated by deriving a subgraph from the establwashed graph using QAOA. These were employed as a local solver and integrated with the Tabu search algorithm for solving large-scale graph-theoretic feature selection (GTFS) problems utilizing limited quantum bit resources. They conducted experiments on 20 datasets and demonstrated that the proposed QAOA_GTFS obtains better results than the full feature set and benchmark algorithms.

Abdulhussein et al. [31] proposed a novel quantum-inspired GA (QIGA) by hybridizing it with one class SVM (OCSVM) for Arabic signature verification. The quantum rotation gate was employed in QIGA to enhance the diversity and convergence properties of the vanilla GA. Moreover, the structure of the model was not predetermined, giving enormous room for the evolutionary algorithm to determine the models' structure by employing the superposition of the Qubit states. The results demonstrated the superiority of the proposed approach in determining the optimal features and outperformed several deep learning features and filter-based approaches. It turned out that QIGA improved the 10-20% error rate of the Arabic signature verification with the benefit of not affecting the computation time.

Ahmad et al. [32] proposed an improved leukemia identification pipeline in white blood cells through an efficient quantum-inspired evolutionary algorithm (QIEA). Initially, deep learning methods were employed to extract the features. In the second stage, QIEA was employed to select an optimal feature subset, thereby reducing the complexity of the model without affecting its accuracy. They employed dwastinct quantum rotation for different individual variables of candidate solutions to improve the exploration of search space. They demonstrated that the proposed methodology was validated on the public dataset of 5000 images of five subtypes of WBCs. Their approach obtained a phenomenal reduction in the feature vector (i.e., 90% of the total features) with a classification accuracy of about 99%. Moreover, it turned out to show a better convergence performance than the classical genetic algorithm and a comparable performance to several exwasting works.

Abdulhussein et al. [33] improved the performance of their proposal [15] by adopting the following modification in their proposal to improve the efficacy over the Arabic handwritten datasets: (i) they generated the final fusion vector by combining the embeddings from various feature extraction techniques. (ii) they also combined GA operators with the quantum superposition principles over the Arabic handwritten datasets.

## 6.2 Variants of Quantum Particle Swarm Optimization

Zhang et al. [34] proposed Multi-class quantum based PSO (MQBPSO) for intrusion feature selection and detection. A *probabilwastic mutation* was adopted to avoid local optima, and a *tabu search table* was used to enlarge the quantum particle swarm's search space and avoid repeated computation. They also employed a quantum rotation gate, where the rotation angle was adjusted according to the best solution based on the random number. Through comparwason experiments with the classical intrusion feature selection methods, they found that a correlation relationship exwasts among network intrusion features. The MQBPSO-based wrapper feature selection demonstrated superiority over classical methods. The efficacy of the proposed approach was tested on the various KDD Cup 99 datasets.

Hamed et al. [35] proposed quantum-inspired PSO (QIPSO) for evolving spiking neural network (ESNN). Using a wrapper approach, their method simultaneously optimized ESNN parameters and relevant features. The main idea of QIPSO was to use a standard PSO function to update the particle position in a Qubit. However, the velocity update was modified to get a new quantum angle, which was then translated into a new probability of the angle, helping to maintain the diversity of the PSO. The results demonstrated that the new algorithm effectively obtained the parameters and identified important features.

Gong et al. [36] proposed Genetic Quantum PSO (GQPSO) for network intrusion detection, combining normalized mutual information between features and a fitness function guided by attribute reduction to realize optimal selection of network data feature subset. Thwas method optimally selects network intrusion features by dwascarding independent and redundant attributes. The efficacy of the proposed approach was studied using various KDD cup 99 datasets. From experimental results, the authors showed that the classification detection rate and detection speed of the GQPSO algorithm were higher than those of the PSO and QPSO algorithms.

Bekri and Govardhan [37] proposed an escalated mediocre agent-based quantum particle swarm optimization (EMA-QPSO). They proposed that the quadratic polynomial technique should be considered for best-fit swarm particles to generate the offspring. These offspring replace the least swarm particles if they obtain higher accuracy. They demonstrated the scalability of the EMA-QPSO in handling high-dimensional datasets. Their approach was validated on the diabetes dataset where EMA-QPSO turned out to be superior to Tabu search and other extant algorithms.

Jin and Jin [38] proposed an improved quantum PSO (IQPSO) for vwasual feature selection (VFS), and an ensemble strategy was employed to improve the FSS performance. They employed a reverse operation, where the better solution, which was away from the particle's position, was often verified and decided by a random number. Here, only 50% of the solutions were verified in the search space away from the current search space. If any of these solutions were found to be efficient, then they replaced the original solutions. Thwas avoided premature convergence and ensures population diversity. The effectiveness of their approach was analyzed over a large image dataset by employing an ensemble classifier, which turned out to be better than the standalone classifiers.

Illiyasu and Fatichah [39] proposed a hybrid method combining quantum PSO and Fuzzy KNN to select optimal feature subsets for detecting cervical cancer. Initially, features were extracted using extant methods, and QPSO was employed to select the optimal feature set. They introduced a quantum mechanical interpretation of the wave function of the particle at the current position. Thier approach demonstrated that, after performing feature selection, the model's accuracy either outperformed or performed as good as the accuracy without feature selection, with a great reduction in the length of the feature subset.

Zouache et al. [40] proposed the FSS method by hybridizing the *firefly algorithm* and PSO algorithm to a condition feature subset and named it QCWASA-FS. They employed two different strategies to guide the exploration phase of their proposed algorithm: (i) The first stage was employed by selecting two solutions randomly, and the cardinality was compwered. If solution 1 has a higher cardinal than solution 2, then it attracts solution 2; (ii) if solution 1 turns out to be better than solution 2, then solution 1 was compwered with the corresponding best ~~and best~~ and gets attracted towards the better position. They compwere the performance of the QCWASA-FS algorithm with the exwasting rough set theory-based feature selection algorithms over 11 different datasets. In most of the datasets, QCWASA-FS turned out to be the best-performing algorithm.

Chaudhari and Agarwal [41] introduced a novel method with quantum particle swarm optimization with elitwast breeding (EBQPSO) on gene data sets. Thwas breeding strategy acts on the elitwasts of the swarm to escape from the likely local optima and helps to perform search efficiently. It improves the search efficiency of QPSO, and the mechanwasm of updating the elitwasts with the new bred individuals with better fitness also provides a more efficient and precwase search guidance for the swarm. They employed a *transposon operator* to maintain the diversity among the solutions where the jumping rate decides the conversion of each particle to a different search space. EBQPSO conswastently improves the result on gene datasets where the approach turned out to be superior to the exwasting methods by yielding higher accuracies.

Wu et al. [42] proposed a two-stage based feature selection to solve FSS. In the first stage, the correlation coefficient measure, i.e., maximal information coefficient (MIC), was employed to filter out the unnecessary features. Here, the weaker the correlation features, the more unlikely a feature could be a redundant one. In the second stage, the filtered features were provided to improve the quantum particle swarm optimization (iBQPSO) algorithm wherein the crossover between two successive bests ~~and best~~ was performed to generate a new individual. Further, they employed Hamming dwastance to decide the probability of attraction towards the local position. iBQPSO algorithm can achieve better classification accuracy than the BQPSO, FCBF, and SVMRFE algorithms on four high-dimensional datasets.

Wu et al. [43] proposed a hybrid improved quantum PSO (HI-BQPSO), which was also another two-stage method with a filtering method and wrapper. Thwas proposal also works in two stages: (i) the MIC was employed to reduce the dimensionality of the data, and (ii) the filtered features were fed to HI-BQPSO for better exploration feature subsets. HI-BQPSO also employs a novel mutation-based operator, wherein the bit of an individual will be flip-flopped. Thwas operation was performed if and only if the fitness of a particular solution was greater than the average fitness of the population. Further, a random toss will be performed to decide on an individual bit. The efficacy of their approach was studied in nine gene expression datasets and 36 UCI datasets to evaluate and compwere the classification accuracy of the HIBQPSO's selected feature subsets against four other algorithms. HI-BQPSO has good overall performance and strong searchability, and it was able to maintain high efficiency with a range of different classifiers.

Jufu et al. [44] proposed chaotic QPSO (CQPSO) to solve feature selection in Mahalanobwas-Taguchi system (MTS). They addressed the slow convergence problem of standard binary particle swarm optimization (BPSO) by introducing logwastic chaotic map-based initialization. They employed reverse transformation of generated chaotic sequence and employed it to improve the diversity and convergence criteria. The efficacy of their proposed approach was proved in terms of better convergence, obtaining efficient solutions, i.e., higher accuracy with fewer cardinal feature subsets in MTS systems settings.

Zhang et al. [45] proposed a framework addressing both the data scarcity and the feature selection. Initially, Wasserstein Generative Adversarial Network (W-GAN) was used to generate the samples. Later, kernel partial least squweres based quantum PSO (KPLS-QPSO) was employed wherein KPLS was used as a filter to remove irrelevant features and QPSO was employed to select optimal feature subset. The results demonstrated that KPLS-QPSO handled the high dimensionality well and obtained higher accuracy results by outperforming the extant approaches for credit rwask assessment.

Agarwal et al. [46] proposed a annealing levy QPSO (ALQPSO) where various principles of quantum mechanics were employed, such as quantum superposition, quantum rotation gate, and entanglement. ALQPSO turned out to be outperforming the corresponding non quantum variant and obtained higher accuracy and convergence speed with less cardinality.

### 6.3 Variants of Quantum Differential Evolution

An elitwast-based Quantum-inspired Differential Evolution (QDE), proposed by Srikrwashna et al. [47], converts the Q-bit to a single variable $\theta$ with the Cartesian coordinates given by sin $\theta$ and cos $\theta$ , where $\theta$ was defined in $[0,2\pi]$. Thwas conversion preserves the information of the $\alpha$ and $\beta$ states in $\theta$. QDE employs an elitwasm principle, preserving elite solutions unchanged throughout the evolutionary process. The elite solution was determined by the accuracy of the population, which in turn improves the non-elite solutions. Elitwasm was introduced from the first iteration and was executed if the accuracy of the corresponding solution was less than 90 percent. Though QDE outperforms its non-quantum variant counterpart in terms of accuracy., it was a non-adaptive variant, necessitating the user to set the parameters.

Ng et al. [48] proposed a quantum DE with SVM and applied it for gene feature selection. Taking inspiration from Srikrwashna et al. [47], they also introduced the elitwasm principle based on the elitwasm threshold. However, they used an adaptive threshold that depends on the best F1 score and the adjusted ideal portion of the features to be selected. Thwas approach outperformed the other variants, such as LR, DT, and ELM. Further, their approach also outperformed several feature selection wrapper methods with a mean F1 score difference of 0.08-0.14.

### 6.4 Variants of Quantum Grasshopper Optimization Algorithm

Wang et al. [49] proposed a quantum variant of the GOA algorithm named the quantum GAO (QGAO) algorithm. They employed a dynamic population strategy to generate solutions from different search spaces, effectively handling catastrophe conditions and preventing the algorithm from becoming trapped at local optimal solutions. A quantum rotation gate was used, where the rotation angle and direction were adaptive and decided based on the positions of the best particle relative to the current particle. Thwas adaptation helped mitigate premature convergence, providing an additional advantage. The algorithm's parameters were adaptive and were decided by the roulette wheel mechanwasm. Their proposal was validated across twenty UCI datasets, demonstrating superior performance compwered to exwasting approaches in the majority of datasets.

### 6.5 Variants of Quantum Whale Optimization Algorithm

Agarwal et al. [50] proposed a quantum whale optimization algorithm (QWOA) where a *quantum rotation gate* serves as a variation operator. The direction and rotation angle were decided by the lookup table. They further introduced modified mutation and crossover operators tailored for quantum-based improvements in exploration capabilities compwered to non-quantum algorithms. These new operators enhances exploration by facilitating shrinking and spiral movement of the whales within QWOA. To validate the effectiveness of their algorithm, they compwered it with the conventional WOA across fourteen publicly available datasets spanning diverse domains.

Kaur et al. [51] proposed a two-phase approach wherea filter measure was employed to weed out the unnecessary features, and thereafter an optimal feature subset was obtained through the nature-inspired wrapper-based feature selection for employing QWOA. The validity of the QWOA was proved by conducting experiments on the publicly available Dwastress Analyswas Interview Corpus Wizard-of-Oz (DAICWOZ) dataset. The two major advantages of thwas approach were (i) low computational complexity in comparwason to traditional wrapper-based evolutionary methods and (ii) easy to implement. The efficacy of QWOA approach reflects the superiority in comparwason to exwasting unimodal and multimodal automated depression detection models.

### 6.6 Variants of Quantum Immune Colonal Algorithm

Soliman and Ressman [52] introduced a quantum bio-inspired estimation of dwastribution algorithm (EDA) for CFS. The proposed algorithm integrates the quantum computing concepts and vaccination process with the immune clonal selection (QVICA). The main objective of introducing quantum concepts was to improve the searchability and decrease the convergence time. Their proposed methodology conswasts of the following phases: (i) all the symbolic data was converted to the numeric value, and then data dwascretization was performed where they employ equal frequency dwascretization (EFD) technique, and (ii) the correlation matrix was constructed, and then the metaheurwastic was employed to search for the optimal combination of features. It was implemented and evaluated using the benchmark intrusion detection dataset KDD Cup99 and compwered with the GA algorithm. The results indicate that the proposed quantum algorithm was efficient in obtaining optimal feature subsets.

### 6.7 Variants of Quantum Moth Flame Optimization Algorithm

Dabba et al. [53] embedded the quantum computing principles into the moth flame optimization algorithm and designed a quantum-inspired version of it. They proposed a two-stage methodology, where minimum redundancy and maximum relevance (mRMR) were employed to filter out the redundant and irrelevant features, and filtered features were fed to the proposed quantum moth flame optimization algorithm (QMFOA) to identify the most important feature subsets. They employed a rotation gate, which drives the search process. The efficacy of their proposal was tested on the thirteen microarray gene datasets and compared with various state-of-the-art methods.

### 6.8 Variants of Quantum Hummingbird Algorithm

The data collected from various interconnected Social Internet of Things (SIoT) systems was massive, demanding robust and efficient processing algorithms, feature extraction, selection, and inference. Elaziz et al. [54] embedded the concepts of quantum computing with the Artificial Hummingbird algorithm (AHA) and employed it for FSS. They employed quantum computing with the motive of improving the exploration capability of the algorithm while discovering feasible regions. The quantum inference principle was employed with a quantum rotation gate, and the angle was decided by a reference table. Their main objective of using the quantum was to enhance the exploration of agents while determining the optimal subset of features. They conducted extensive experiments on eighteen different datasets collected from the UCI repository.

### 6.9 Variants of Quantum Ant Bee Colony

Zhong et al. [55] proposed a self-adaptive quantum equilibrium optimizer (EO) with an artificial bee colony and named it SQEOABC. In their work, they assumed that the quantum behaviour of each particle was determined by the wave function. Hence, QEO updates the particles by utilizing the whole population information and introducing the average best position of particles. The execution of ABC or QEO was decided by their respective coefficients. These coefficients were designed by considering equal preference for both exploration and exploitation. Their proposed self-adaptive operator was introduced to enhance the convergence speed of the algorithms. Their proposal was validated on 25 datasets collected from the UCI repository, where the proposed approach turned out to be superior in obtaining higher accuracy with a great reduction in cardinality. In addition, they also analyzed COVID-19 data to illustrate the effectiveness of their approach. Their results concluded that SQEOABC outperformed the standard EO and several other variants of EO.

### 6.10 Variants of Quantum Squirrel inspired Algorithm

Ghosh et al. [56] proposed a squirrel-inspired algorithm for FSS by exploiting the benefits of the quantum computing principle for the Squirrel search algorithm. We were awwere of the concept of "end of winter" in the context of the squirrel algorithm to accelerate the convergence rate of the algorithm. The authors have introduced a modified version of thwas concept to accelerate the convergence in quantum settings. Thwas affected the algorithm's performance while achieving higher classification accuracy with a large reduction in feature dimensions. Further, a relief score was employed to rank the features of the genes, which were highly effective in classifying the genes. Further, the proposed quantum-inspired algorithm can achieve 100 percent accuracy in most datasets, with a substantial reduction in the feature subsets compwered to several traditional algorithms.

### 6.11 Variants of Quantum Owl Search Algorithm

Mandal et al. [57] proposed a quantum-inspired owl search algorithm (QIOSA) where a quantum rotation gate was employed to accelerate the search towards an optimal set of features. They employed a quantum rotation gate, where the rotation angle was defined by the intensity difference between the owl and the prey's binary location. However, the direction of the rotation angle was defined by the alpha and beta of a Qubit state. Through results, it was demonstrated that their proposed outperformed several exwasting algorithms, such as GA and PSO, in twelve publicly available benchmark datasets with a difference of 1-3%.

### 6.12 Variants of Quantum Avian Navigation

Nadimi-Shahraki et al. [58] developed an efficient binary version of the quantum-based avian navigation optimizer algorithm (QANA) named BQANA. They utilized the scalability of the QANA to effectively select the optimal feature subset from high-dimensional medical datasets using two different approaches using two different approaches. (i) The efficacy of various transfer functions was examined, including S-shaped, V-shaped, U-shaped, Z-shaped, and quadratic transfer functions, to map the continuous solutions of the canonical QANA to binary ones. (ii) The binary version of QANA was designed and assessed using different medical datasets. The results demonstrated that the binary version, i.e., BQANA, yielded superior results compared to other binary versions of the QANA in finding the optimal feature subset in the medical datasets.

Fatahi et al. [59] proposed an improved binary QANA (IBQANA) with a Hybrid Binary Operator (HBO) to convert continuous values within any range into binary solutions. Moreover, a Distance-based Binary Search Strategy (DBSS) was also introduced, which was a two-phase search strategy that enhanced the performance of inferior search agents and accelerated convergence. The exploration and exploitation were managed by an adaptive probability function, preventing the algorithm from getting stuck at local optima. The effectiveness of applying HBO was compared with five distinct transfer function families and thresholding on 12 medical datasets, with feature numbers ranging from 8 to 10,509. Furthermore, IBQANA was utilized to detect COVID-19, which outperformed several existing methods on COVID-19 and 11 other medical datasets.

### 6.13 Variants of Quantum Dwarf mongoose algorithm

Almutairi et al. [60] proposed an inspired dwarf mongoose algorithm (QDMO-EDLID) for intrusion detection in cyber physical systems (CPS). It works in two stages. In the first stage, features were fed to the QDMO algorithm to identify important feature subsets. Such feature subsets were passed to an ensemble of Convolution Residual Networks (CRN), Deep Belief Networks (DBN), and Deep Autoencoder (DAE) techniques to identify the presence of intrusions in CPS space. In QDMO, they employed a quantum rotation gate where the rotation angle was adjusted according to the reference table, which considers the fitness value of the best solution. Their proposed approach, the QDMO-EDLID technique, was tested on various benchmark intrusion detection datasets wherein it turned out to be better than the standalone algorithms Convolution Residual Networks (CRN), Deep Belief Networks (DBN), and Deep Autoencoder (DAE) techniques by more than 1-6% with a reduction in complexity.

## 6.14 Variants of Quantum Honey Badger Algorithm

Alshahri et al. [61] employed a 2D Chaotic Hénon map to adjust the parameters of the modified version of the honey badger algorithm (HBA) and named it quantum chaotic HBA (QCHBA). A chaotic map was introduced to improve the HBA algorithm's exploration capability. The results demonstrated that the proposed QCHBA outperformed several exwasting methods and turned out to be superior in obtaining higher accuracy in the majority of the datasets.

## 6.15 Variants of Quantum Fruit Fly Algorithm

Nijaguna et al. [62] proposed the Quantum Fruit Fly Algorithm (QFFA) technique for feature selection to improve the effectiveness of classification in medical diagnoses. The QFFA technique applies Archimedes' spiral to increase the exploitation of the model, enabling it to select unique features for classification. Here, the distance between the current and the best position was calculated, and then the new offspring were generated by using this information. This kind of mechanism enables the spiral way of searching for the best solutions for the Fruit Fly algorithm, which helps to overcome the local optima trap and increases exploitation. It was demonstrated through results that the QFFA model achieved better performance metrics, such as sensitivity (99.26 %), and accuracy (99.04%), when compared to several other extant deep learning techniques such as Deep 1D-CNN and GA-Decision tree models when applied to medical datasets. Additionally, there was a significant reduction in the percentage of selected features, while maintaining a high level of accuracy.

## 6.16 Variants of Quantum Grey Wolf Algorithm

Zhao et al. [63] proposed quantum computing and uncertain symmetry rough set and named it QCGWORS. The rotation gate was employed in QCGWORS, where the rotation angle was decided by a reference table, as in several previous works. In thwas algorithm, the relevance of the feature subset was calculated by using uncertain symmetry to see the importance of the selected feature subset. The comparative analyswas was conducted between its corresponding non-quantum grey wolf algorithm wherein a few datasets, the classification accuracy was improved up to 20%, which was so phenomenal.

El-ashry et al. [64] proposed EQI-GWO, where the wolf position was a binary value and was converted using the *sigmoid* function. The positions of the wolves depend on both the Qubit associate and the corresponding rotation angle, which was decided by two random probabilwastic random values. It was demonstrated through results that their rotation angle gate-based algorithm obtained superior results than the corresponding non-quantum GWO variant, i.e., BGWO. Thwas consequently resulted in higher accuracy with an effective selection of features.

## 6.17 Variants of Quantum Gravitational Search Algorithm

Han et al. [65] proposed QBGSA, a quantum variant of the gravitational search algorithm (GSA). Here, the rotation angle was added in place of the random variables to update the velocity of a particle. The rotation angle was adaptive and was not decided by any predetermined table. Further, they employed KNN as the classifier. They demonstrated that the number of features obtained by their approach was low, yet accuracy was well maintained. Further, their approach has shown the ability to escape from the local optimum and was more robust than binary PSO and GA algorithms.

Barani et al. [66] proposed BQIGSA, a binary quantum variant of the GSA algorithm. Here, the rotation angle was also added in place of the random variables. In quantum computing, the angular velocity determines the value of the movement towards 1 (selection of the feature) or 0 (exclusion of the feature). Here, they proposed an adaptive rotation angle based on the angular velocity, which was decided based on the deviation between the best solution and the current solution. The results empirically demonstrated that thwas approach obtained the best precwasion and recall with a smaller number of features than several extant approaches in several datasets. Further, thwas approach also outperformed the QBGSA [1] algorithm, showcasing the effectiveness of the inclusion of angular velocity dwasplacement in the rotation angle.

Noormahammadi and Dowlatshahi [67] proposed a memetic algorithm for multi-label datasets that offers a search algorithm with BQIGSA. Thwas local search algorithm conswasts of the prior and posterior knowledge vectors that help in deciding the inclusion or exclusion of irrelevant and redundant features. The rotation angle was also an adaptive one and was decided based on the deviation between the best solution and the current solution. In their local search algorithm, the best solution was used to generate eight different neighbors as follows: (i) the first four neighbors were generated by flipping four bits from zero to one, and (ii) the second four neighbors were generated by flipping four bits from one to zero. These features were selected based on the highest importance. If any of the neighbors turn out to be better than the best solution, then it will replace the best solution. Further, the total knowledge vector was also calculated as an additive combination of the prior and posterior knowledge vectors. The results gave empirical evidence that their approach outscored the extant approaches in three of five datasets.

## 6.18 Variants of Quantum Immune clone optimization algorithm

Eluri et al. [68] proposed a quantum-inspired immune clone optimization technique for optimal feature selection in a classification task. The model mainly comprwases three main phases: (1) Gene expression data collection and feature extraction: gene data was collected, and features were extracted; (2) Optimal feature selection: quantum-inspired immune clone optimization algorithm (QICO) was employed to select the optimal feature subset, which was carried out by a recurrent neural network (RNN) and (3) Classification: The selected features were then classified using RNN. Initially, five benchmark gene expression datasets were collected. Once the relevant features were selected, the classification was performed by a deep learning model called recurrent neural network (RNN). Finally, the experimental analyswas demonstrated that the proposed QICO-based feature selection model QICO-RNN was 3.2% better than RNN in terms of the accuracy of the model.

## 6.19 Variants of Quantum Shuffled Frog Leaping Algorithm

Ding et al. [69] proposed a quantum variant of the shuffled frog leaping algorithm and named it QSFLAR. Here, they employed multi-state quantum bits notation to denote evolutionary frogs to improve the diversity and convergence of the algorithm. The states of the quantum-bit were initialized with the same probability. They employed two different quantum gates, namely, (i) quantum rotation gate in which the angle of rotation was decided by the lookup table, and (ii) quantum NOT gate to mimic the behavior of the quantum mutation operation. Further, they divided the total feature set into different subsets and employed the co-evolving principle where the best feature subsets were obtained based on the co-shwered information among the elitwast frog in each of the subsets. They demonstrated that QSFLAR was the best performing algorithm by selecting fewer cardinal subsets. Further, it maintained a balance between exploration and exploitation, thereby maintaining diversity and convergence to obtain the global optimal solution.

**Table 4: Summary of the QIEA based FSS papers**

| Algorithm | Objective | Single/ Multi-objective | S/P | Classifier in the Wrapper | Application | A | Data partitioning | Environment | Encoding Scheme | Gates |
|---|---|---|---|---|---|---|---|---|---|---|
| Genetic Algorithm (GA) | | | | | | | | | | |
| Quantum GA (QGA) [14] | Maintain population diversity and avoid selection pressure | Single (Error rate) | S | Neural Network (NN) | Radar emitter signals | | Hold out (H) | Classical | Q | Rotation gate |
| Novel Quantum GA (NQGA) [15] | Enhance search capability and premature convergence | Single | S | NN | Radar emitter signals | ✓ | H | Classical | Q | Rotation gate |
| Quantum-inspired clone GA(QCGA) [16] | Avoid premature convergence | Single | S | K Newerest Neighbour (KNN) | Text categorization | | H | C | Q | NO |
| QGA [17] | To select effective feature subsets in text classification | Single | S | NB | Text classification | | H | C | Q | Rotation gate and NOT gate |
| Genetic Quantum Algorithm with SVM ($GQA_{SVM}$) [18] | To address difficulties in high dimensional datasets | Single | S | SVM | Microarray data | | H | C | Q | Rotation gate |
| ClusterQGA [19] | To explore clustering algorithms to select a small set of features | Single | S | LDC, SVM, KNN | Microarray data | | H | C | Q | Rotation gate |
| QGA [20] | To select sensitive features in fault simulation test | Single | S | NN | Fault simulation test | | H | C | Q | Rotation gate |
| Quantum GA [21] | Enhance classification accuracy | Single | S | NN | Character recognition | | H | C | Q | Rotation gate |
| Quantum-inspired GA (QiEA) [22] | To select the most relevant features | Single | S | Multi-layer perceptron (MLP) | Brain computer interface (BCI) | | H | C | Q | Rotation gate |

**Table 4: Summary of the QIEA based FSS papers (contd.)**

| Algorithm | Objective | Single/ Multi-objective | S/P | Classifier in the Wrapper | Application | A | Data partitioning | Environment | Encoding Scheme | Gates |
|---|---|---|---|---|---|---|---|---|---|---|
| Genetic Algorithm (GA) | | | | | | | | | | |
| Principal component analysis-Quantum evolutionary algorithm (PCA-QEA) [23] | Concentrate on the search spaces where there is less certainty about whether to select features or not | Single | S | Cascade forward neural network (CFNN); NN; Fuzzy neural network (FNN); Generalized regular regression (GRNN); KNN; Linear Discriminant Classifier (LDC); Naive Bayes (NB); SVM | Identifying fillers in the text | | H | C | Q | NO |
| Chaotic Quantum-inspired evolutionary algorithms (CQiEA) [24] | To enhance the convergence rate and speed of the EA | Single | S | MLP | BCI | | H | C | Q | Chaotic Rotation gate |
| Quantum-inspired Genetic Algorithm (QIGA) [25] | To select minimum features with high accuracy | Single | S | SVM | Microarray datasets | | 10-fold | C | Q | Rotation gate |
| Hybrid quantum feature selection algorithm (HQFSA) [26] | To explore graph benefits from the quantum perspective | Single | S | SVM-linear | Generic | | 10-fold | C | Graph | CNOT gate |

**Table 4: Summary of the QIEA based FSS papers (contd.)**

| Algorithm | Objective | Single/ Multi-objective | S/P | Classifier in the Wrapper | Application | A | Data partitioning | Environment | Encoding Scheme | Gates |
|---|---|---|---|---|---|---|---|---|---|---|
| Genetic Algorithm (GA) | | | | | | | | | | |
| Principal component analysis-Quantum evolutionary algorithm (PCA-QEA) [23] | Concentrate on the search spaces where there is less certainty about whether to select features or not | Single | S | Cascade forward neural network (CFNN); NN; Fuzzy neural network (FNN); Generalized regular regression (GRNN); KNN; Linear Discriminant Classifier (LDC); Naive Bayes (NB); SVM | Identifying fillers in the text | | H | C | Q | NO |
| Chaotic Quantum-inspired evolutionary algorithms (CQiEA) [24] | To enhance the convergence rate and speed of the EA | Single | S | MLP | BCI | | H | C | Q | Chaotic Rotation gate |
| Quantum-inspired Genetic Algorithm (QIGA) [25] | To select minimum features with high accuracy | Single | S | SVM | Microarray datasets | | 10-fold | C | Q | Rotation gate |
| Hybrid quantum feature selection algorithm (HQFSA) [26] | To explore graph benefits from the quantum perspective | Single | S | SVM-linear | Generic | | 10-fold | C | Graph | CNOT gate |

**Table 4: Summary of the QIEA based FSS papers (contd.)**

| Algorithm | Objective | Single/ Multi-objective | S/P | Classifier in the Wrapper | Application | A | Data partitioning | Environment | Encoding Scheme | Gates |
|---|---|---|---|---|---|---|---|---|---|---|
| Genetic Algorithm (GA) | | | | | | | | | | |
| QGA [27] | To reduce cost and computation time to build motor imagery | Single | S | SVM | Motor imagery | | 10-fold | C | Q | Rotation gate |
| Quantum GA (QGA) [28] | To perform both feature selection and architecture optimization | Single | S | SVM | Predictive emissions monitoring system (PEMS) | ✓ | H | C | Q | Rotation gate |
| Normalized mutual information feature selection and cooperative evolution of multiple operators based on adaptive parallel quantum GA (NMIFS MOP-AQGA) [29] | To address the problems that the IDS has lower detection speed, less adaptability, lower detection accuracy | Single | P | SVM | Intrusion detection system (IDS) | ✓ | H | C | Q | Rotation gate |
| Quantum approximate optimization algorithm (QQOA) [30] | To solve graph theoretic problems | Single | S | MCFS, DSFFC, U-FAM | Generic | | Stratified 10-fold | C (qiskit) | Graph | Hadamard/ NOT-gate/ Rotation gate |

**Table 4: Summary of the QIEA based FSS papers (contd.)**

| Algorithm | Objective | Single/ Multi-objective | S/P | Classifier in the Wrapper | Application | A | Data partitioning | Environment | Encoding Scheme | Gates |
|---|---|---|---|---|---|---|---|---|---|---|
| Quantum-inspired GA (QIGA) [31] | To improve exploration and exploitation | Single | S | One class-SVM (OC-SMV) | Arabic signature verification | | H | C | Q | Rotation gate |
| QIEA [32] | To enhance exploration capability | Single | S | Analyzed with four different classifiers: KNN, SVM, Decision Tree (DT), NN | Leukemia detection | | H | C | Q | Rotation gate |
| Improved QIGA (IQIGA) [33] | To address computational complexity and obtain early convergence | Single | S | OC-SVM | Arabic signature authentication | | H | C | Q | Rotation gate |
| Particle Swarm Optimization (PSO) | | | | | | | | | | |
| Quantum Binary PSO (QBPSO) [34] | To avoid local optima | Single | S | SVM | Intrusion detection | | H | C | Q | Rotation gate |
| Evolving spiking NN- Quantum-inspired PSO (ESNN-QIPSO) [35] | To perform both feature and parameter optimization | Single | S | Spiking Neural Network (SNN) | Generic | | H | C | Q | Rotation gate |
| Genetic Quantum PSO (GQPSO) [36] | To improve the rate of intrusion detection and classification rate | Single | S | SVM | Intrusion detection | | H | C | B | None |
| Improved QPSO (IQPSO) [38] | To improve population diversity | Single | S | Ensemble classifier | Automatic image annotation (AIA) | | H | C | B | None |

**Table 4: Summary of the QIEA based FSS papers ( contd ..)**

| Algorithm | Objective | Single/ Multi-objective | S/P | Classifier in the Wrapper | Application | A | Data partitioning | Environment | Encoding Scheme | Gates |
|---|---|---|---|---|---|---|---|---|---|---|
| Improved quantum PSO (IQPSO) [39] | | Single | S | KNN; LDA; SVM | Image | | H | C | B | None |
| Quantum behaved PSO (QPSO) [40] | To accelerate convergence | Single | S | Fuzzy KNN | Cervical cancer detection | | 5-fold | C | B | None |
| Quantum computation concepts and rough set (QCSIA-FS) [41] | For effective exploration | Single | S | KNN, CART | Generic (Studied on 11 datasets) | | H | C | B | None |
| PSO with elitism breeding (EBQPSO) [42] | Consideration of elitism for deep exploration | Single | S | SVM, J48, NN | Cancer classification | | H | C | B | None |
| Improved Binary QPSO (IBQPSO) [43] | To remove weak correlation features | Single | S | SVM | Generic (Studied on 4 datasets) | | 5-fold | C | B | None |
| Hybrid improved-quantum behaviour PSO (HI-QBPSO) [44] | Combines filtering method with the wrapper | Single | S | SVM | Generic (studied on 36 datasets) | | H | C | B | None |
| Chaos quantum behaviour PSO (CQPSO) [45] | Effectively enhance the iterative speed and optimization precision of the particles | Single | S | | Steel plate fault dataset | | H | C | B | None |
| Kernel partial least squweres quantum PSO (KPLS-QPSO) [46] | Better convergence | Single | S | SVM; KNN; LogR; AdaBoost | Generic | | H | C | B | None |

**Table 4: Summary of the QIEA based FSS papers (contd ..)**

| Algorithm | Objective | Single/ Multi objective | S/P | Classifier in the Wrapper | Application | A | Data partitioning | Environment | Encoding Scheme | Gates |
|---|---|---|---|---|---|---|---|---|---|---|
| Differential Evolution | | | | | | | | | | |
| Quantum-inspired DE [47] | For faster convergence | Single | S | LR | Generic (studied on three datasets) | | 10-fold | C | B | None |
| Quantum-inspired DE (QDE) [48] | To improve the convergence rate | Single | S | SVM | scRNA-seq data | | 5-fold | C | B | None |
| Grasshopper optimization algorithm | | | | | | | | | | |
| DQBGOA_MR [49] | To avoid premature convergence and avoid getting entrapped in a local optimal solution | Single | S | J48; Naïve Bayes | Generic (studied on 20 datasets) | ✓ | 10 Fold | Classical | Q | Rotation gate (Adaptive) |
| Whale optimization algorithm | | | | | | | | | | |
| Quantum whale optimization algorithm (QWOA) [50] | Enhances diversification and convergence | Single | S | KNN, LDA, SVM, C4.5 | Generic (studied on 14 datasets) | | H | C | Q | Rotation gate |
| Quantum-based whale optimization algorithm [51] | To enhance population diversity | Single | S | KNN; LDA; SVM | Distress analysis | | H | C | Q | Rotation gate |
| Immune colony optimization algorithm | | | | | | | | | | |
| Quantum, vacation process with immune colony algorithm (QVICA) [52] | To improve the search performance and reduce computation time | Single | S | | Intrusion detection | | H | C | Q | |

**Table 4: Summary of the QIEA based FSS papers (contd.)**

| **Algorithm** | **Objective** | **Single/ Multi objective** | **S/P** | **Classifier in the Wrapper** | **Application** | **A** | **Data partitioning** | **Environment** | **Encoding Scheme** | **Gates** |
|---|---|---|---|---|---|---|---|---|---|---|
| Mothflame optimization algorithm | | | | | | | | | | |
| Quantum moth flame optimization algorithm (QMFOA) [53] | To guarantee a good trade-off between exploration and exploitation | Single | S | SVM | Microarray data | | Leave one out CV | C | Q | Rotation gate |
| Hummingbird Algorithm | | | | | | | | | | |
| Quantum artificial hummingbird (QAHA) [54] | To improve exploration capability while discovering feasible regions | Single | S | KNN | Generic (studied on 18 datasets) | | H | C | Q | Rotation gate |
| Ant bee colony | | | | | | | | | | |
| Self-adaptive quantum equilibrium optimizer with ABC (SQEOABC) [55] | To improve the convergence | Single | S | KNN | Covid-19 | ✓ | H | C | Q | None |
| Squirrel inspired algorithm | | | | | | | | | | |
| Quantum squirrel inspired algorithm [56] | To maintain the balance between exploration capability without affecting convergence | Single | S | KNN | Medical datasets | ✓ | 5-fold | C | Q | Rotation gate |
| Owl search Algorithm | | | | | | | | | | |
| Quantum-inspired owl search algorithm (QIOSA) [57] | To accelerate the search toward optimal feature subsets | Single | S | KNN | Generic (Studied on 12 datasets) | | H | C | Q | Rotation gate and Hadamard gate |

**Table 4: Summary of the QIEA based FSS papers (contd.)**

| Algorithm | Objective | Single/ Multi objective | S/P | Classifier in the Wrapper | Application | A | Data partitioning | Environment | Encoding Scheme | Gates |
|---|---|---|---|---|---|---|---|---|---|---|
| Avian Navigation algorithm | | | | | | | | | | |
| Quantum-based avian navigation algorithm (QANA) [58] | To enhance convergence speed | Single | S | KNN | Generic (Studied on 10 medical datasets) | | H | C | Q | None |
| Improved binary QANA (IBQANA) [59] | To improve the exploration phase | Single | S | KNN | Covid-19 | | 10-fold | C | Q | None |
| Dwarf mongoose Algorithm | | | | | | | | | | |
| Quantum dwarf mongoose optimization with ensemble deep learning based intrusion detection (QDMO-EDLID) [60] | To recognize the presence of intrusions via FSS | Single | S | Convolutional residual networks (CRN); Deep Belief Networks (DBN); Deep Auto Encoder (DAE) | Intrusion detection | | H | C | Q | Rotation gate |

**Table 4: Summary of the QIEA based FSS papers (contd.)**

| Algorithm | Objective | Single/ Multi objective | S/P | Classifier in the Wrapper | Application | A | Data partitioning | Environment | Encoding Scheme | Gates |
|---|---|---|---|---|---|---|---|---|---|---|
| Honey badger algorithm | | | | | | | | | | |
| Quantum chaotic HBA [61] | To maintain the balance between exploration and exploitation | Single | S | KNN | Generic (studied on 18 datasets) | | H | C | B | None |
| Fruit fly algorithm | | | | | | | | | | |
| Quantum fruit fly algorithm [62] | Avoid local optima traps and enhance exploitation | Single | S | SVM | Medical diagnosis and cauld also be applied to generic datasets (studied on five datasets)A | | H | C | B | None |
| Greywolf Optimizer | | | | | | | | | | |
| Enhanced quantum-inspired binary grey wolf algorithm (EQI-GWO) [63] | To enhance search capability and improve performance | Single | S | KNN | Generic (studied on four datasets) | | H | C | Q | Rotation gate |
| Grey wolf optimizer based on quantum computing and uncertain symmetry roughest theory (QCGWORS) [64] | To maintain a good balance between exploring global and local features | Single | S | KNN; RF | Generic (studied on thirteen datasets) | | 10-fold | C | Q | Rotation gate |

## 7. Basic building blocks of the works reviewed

In this section, we synthesize the works reviewed by summarizing the components of the proposed approaches that play an important role in the performance, viz., (i) objective function and (ii) variants of the rotation gate employed in the study. We specifically chose the rotation gate among the popular quantum gates since it is the most often employed quantum gate in the literature so far.

### 7.1 Popular Objective Functions

The evolutionary algorithm's performance is mainly dependent on the employed objective function. It acts as a guiding factor in determining the global optimal or near-global optimal solutions. It sets goals such as maximizing/minimizing fitness scores. In the literature, we identified, based on the type of the problem, different objective functions, which were presented in Table 5. In summary, the following were the important observations:

- Accuracy is the most popular metric that is employed.
- Many additive weighted combinations of accuracy and cardinality of the selected feature subset have been thoroughly employed in the literature.
- In a few works, an error, that is, a misclassification rate, is employed, which is another form of accuracy.

**Table 5: Various types of objective functions employed in the literature**

| Reference | Min.( ↓) / Max. (↑) | Objective Function |
|---|---|---|
| [14] | ↓ | $$d + \sum_{i=1}^{n-1} \sum_{j=i+1}^{n} q_{bij}$$ where $d$ is the cardinality of the feature vector and $q_{bij} = (b_1 q_{ij}^1)$ & $(b_{22} q_{ij}^2)$ ..& $(b_m q_{ij}^m)$ ; where & is the AND operation, $q_{bij}$ $(i, j = 1,2,..,n)$ is the element of $i^{th}$ row and $j^{th}$ column in matrix $Q_b$ which is the class separability reduction matrix of an individual in the population. |
| [15] | ↑ | $$\text{Between} - \text{class separability}, S_{ij} = \frac{D_{ij}}{C_{ii} + C_{jj}}$$ where i and j were the indices of two different classes, and $$C_{ii} = \max\left\{\left[\frac{1}{M_i^q} + \sum_{k=1}^{M_i^q} \left\| x_{ik}^q - E(X_i^q)\right\|^p\right]^{\frac{1}{p}}\right\}$$ where q=1,2,…, $N$ is the number of features, $M_i^q$ is the number of samples of $q^{th}$ feature of i^th class, $x_{ik}^q$ is the $k^{th}$ sample value of q^th feature of $i^{th}$ class, $X_i^q = [x_{i1}^q, x_{i2}^q, \dots, x_{iM_i^q}^q]$, $E(X_i^q)$ is the expectation of $X_i^q$, and p (p>= 1) is an integer. $$D_{ij} = \min\{ \| E(X_i^q) - E(X_j^q) \| \}$$ where the minimum distance between $i^{th}$ and $j^{th}$ classes. |
| [16] | ↑ | $$accuracy(x) - \frac{F(x) \times cost(x)}{accuracy(x) + 1} + cost_{max}$$ Here $x$ is the feature subset, $accuracy(x)$ is the test accuracy, cost(x) is the sum of measurement costs of the feature subset represented by $x$, $cost_{max}$ is the sum of the costs associated with all of the features. |
| [18] | ↓ | $$\alpha_1 \times \frac{100}{accuracy} + \alpha_2 \times f_s$$ Here f_s is the total number of selected features |

**Table 5: Various types of objective functions employed in the literature (contd.)**

| Reference | Min./ Max. | Objective Function |
|---|---|---|
| [19] | ↑ | $$Accuracy(z) + \frac{1}{S * m * N(z)}$$ Here, Accuracy(z) is the accuracy of the solution z, S is the total number of test samples, m is the number of input features, N(z) is the size of the selected feature subset |
| [20] | ↑ | $$1 - \frac{1}{s} num \sum_{k=0}^{n} \mid y'_p \neq t_p \mid$$ Here $y'_p$ and $t_p$ were the predictive and the true values of the initiating prediction sample, and s is the number of initiating prediction samples. |
| [21] | ↑ | $$\left[\frac{number\ of\ incorrectly\ classified}{Total\ number\ of\ test\ samples}\right]^{-1}$$ |
| [17], [22 - 24], [27] [31], [33 - 35], [37 - 38], [41], [47], [52], [60],[62],[66 - 69] | ↑ | Accuracy |
| [25], [26] | ↓ | $$\rho = \frac{Cov(x,y)}{\sigma_x * \sigma_y}$$ Here Cov is the covariance between two variables and \sigma is the standard deviation of x and y variables. |
| [28] | ↓ | $$\left[\ \exp(a_1\gamma^2) + a_2\left(1 - \frac{\gamma_e}{\gamma}\right)\right][\ \rho_1\ E_{test} +\ \rho_2\ E_{train}]$$ Where $\gamma$ is the total number of parameters., $\gamma_e$ is the total number of effective parameters, $E_{test}$ and $E_{train}$ were errors exhibited over testing and training datasets, $a_1$ and $a_2$ were the weighting factors chosen based on trial and error method; $\rho_1$ and $\rho_2$ were chosen from [0,1] to obtain a weighted average of the testing and training errors. |
| [29], [30] | ↑ | $$NMI(F_i, F_j) = \frac{I(F_i, F_j)}{\left(\sqrt{H(F_i)\ H(F_j)}\right)}$$ Where NMI is the normalized mutual information function, I(.) is the information gain and H(.) is the entropy function. |
| [32], [61], [64], [65] | ↓ | $$\alpha_1 \times error\ +\ \alpha_2\ \times \left(\frac{f_s}{f_t}\right)$$ where $\alpha_1$ and $\alpha_2$ were the weights chosen, error = 1-accuracy, $f_s$ is the features selected and $f_t$ is the total number of features. |

**Table 5: Various types of objective functions employed in the literature (contd.)**

| | | |
|---|---|---|
| [36] | ↓ | $$\frac{\Sigma_j SU(X_i, C)}{\left(\sqrt{\Sigma_i \Sigma_j SU(X_i) SU(X_j)}\right)}$$ Where SU is the symmetric uncertainty, C is the class attribute, $X_i$ is the i$^{th}$ feature |
| [39], [51] | ↑ | F1-score |
| [40], [49], [63] | ↓ | $$\alpha_1 \times \gamma_W(D) + \alpha_2 \times \frac{\lvert W - O\rvert}{\lvert W\rvert}$$ where $\gamma_W(D)$ is the dependence of conditional feature subset W with respect to the feature D. \|O\| represents the subset O of the selected conditional features, \|W\| represents the cardinality of the conditional features. |
| [42] | ↑ | $$\alpha_1 \times (1 - Accuracy) + \alpha_2 \times \left(\frac{f_s * mic_f}{1 + e^{mic_c}}\right)$$ Here $f_s$ is the number of selected features, mic_f is the maximal information coefficient among features, $mic_c$ is the maximal information coefficient between features and class. MIC(D) = max { M(D)x,y} M(D) = I * (D,x,y)/ log min{x,y} I*(D,x,y) = max I (D\|G) where D is a real pair, x and u were positive integers. |
| [43] | ↑ | (1+$\theta$^2) * Acc * norm(F_num)/ $\theta$^2 * Acc + norm(F_num) where norm(F_num)= 1-Fnum/D |
| [44] | ↓ | Error=1-accuracy |
| [45], [58], [59] | ↓ | $$\alpha_1 \times accuracy + \alpha_2 \times \frac{N_{selected}}{N_{tol}}$$ |
| [46], [60] | ↓ | $$\alpha_1 \times error\ rate + \alpha_2 \times \frac{N_{selected}}{N_{tol}}$$ |
| [48] | ↑ | $$F1 + \left[1 - \frac{N_{selected}}{N_{tol}}\right]$$ |
| [50] | ↓ | $$\alpha_1 \times error + \alpha_2 \times N_{selected}$$ |
| [53], [57] | ↑ | $$\alpha_1 \times accuracy + \alpha_2 \times \frac{N_{tol} - N_{selected}}{N_{tol}}$$ |
| [54], [55] | ↓ | $$\alpha_1 \times error + \alpha_2 \times N_{\frac{selected}{N_{tol}}}$$ |
| [56] | ↓ | $$\alpha_1 \times error + \alpha_2 \times N_{\frac{s}{\alpha_1 + \alpha_2}}$$ $\alpha_1$ = 1000; $\alpha_2$=1 |
| [60] | ↑ | Correlation |

## 7.2 Types of Rotation gates

The most popular gate in the QIEA-based FSS is the rotation gate. We noticed that eighteen different rotation gates were employed in the literature. There were two different components that direct the convergence, and the quality of the solution: (i) Sign denotes the direction to which the Qubits need to be rotated, and (ii) magnitude denotes the angle of the rotation that needs to be employed. We noticed that there were 18 different ways to determine the rotation angle and the direction. Most of the works incorporated the static way of employing the rotation gate, which was determined through the lookup table. In the literature, several lookup tables were noticed, and they were presented in Table 6-Table 13.

### 7.2.1 Static Rotation gates

#### 7.2.1.1 Rotation gate type-SI

In the research works [14-15], the following type of rotation gate is employed as given in Eq. 6, presented in Table 6.

$$\theta = k \times f(\alpha_i, \beta_i) \tag{6}$$

where $k$ is a coefficient whose value influences the speed of convergence and $f(\alpha_i, \beta_i)$ is obtained from the lookup table (refer to Table 6, Eq. 6).
In Table 6, there were four different conditions present where the f($\alpha_i$,$\beta_i$) determine the sign of the rotation. Further, the rotation angle (whether clockwise or anti-clockwise) is based on the product of a certain Qubit's ~~alpha~~ $\alpha$ and ~~beta~~ $\beta$ states. The final angle is determined through a user-defined parameter, $k$.

**Table 6: Lookup table for the rotation table type-SI**

| $d_1 > 0$ | $d_2 > 0$ | f($\alpha_i$,$\beta_i$) | |
|---|---|---|---|
| | | $\lvert\zeta1\rvert > \lvert\zeta2\rvert$ | $\lvert\zeta1\rvert < \lvert\zeta2\rvert$ |
| True | True | +1 | -1 |
| True | False | -1 | +1 |
| False | True | -1 | +1 |
| False | False | +1 | -1 |

Here, $d_1 = \alpha_1 \cdot \beta_1$, $\zeta1$= arctan($\beta_1/\alpha_1$) where $\alpha_1$, $\beta_1$ is the probability amplitude of the best solution and $d_2 = \alpha_2 \cdot \beta_2$, $\zeta2$= arctan($\beta_2/\alpha_2$) where $\alpha_2$, $\beta_2$ is the probability amplitude of the current solution.

#### 7.2.1.2 Rotation gate type-SII

Unlike the user-defined magnitude as employed in the rotation gate type-I (refer to section 6.2.1), works in [18],[19],[27],[50] presented rotation gate type-SII (refer to Table 7). Here, the binary encoded vector of the best solution is taken into consideration while employing both the sign and the magnitude. Further, a fixed set of magnitudes has been chosen, and the value of magnitude is determined based on the comparison of the fitness value of the current solution and the best solution. Overall, there were eight different conditions taken into consideration while determining the magnitude and the sign of the rotation.

**Table 7: Lookup table for the rotation gate table type-SII**

| $X_i$ | $b_i$ | F(X) <= F(b) | $\Delta\ \theta_i$ | s($\alpha_i$,$\beta_i$)) | | | |
|---|---|---|---|---|---|---|---|
| | | | | $\alpha_i\beta_i$> 0 | $\alpha_i\beta_i$< 0 | $\alpha_i$=0 | $\beta_i$=0 |
| 0 | 0 | 0 | 0.003 $\pi$ | -1 | 1 | +/- 1 | 0 |
| 0 | 0 | 1 | 0.02 $\pi$ | -1 | 1 | +/- 1 | 0 |
| 0 | 1 | 0 | 0.008 $\pi$ | 1 | -1 | 0 | 1 |
| 0 | 1 | 1 | 0.02 $\pi$ | -1 | 1 | +/- 1 | 0 |
| 1 | 0 | 0 | 0 | 0 | 0 | 0 | 0 |
| 1 | 0 | 1 | 0.035 $\pi$ | 1 | -1 | 0 | +/- 1 |
| 1 | 0 | 0 | 0 | 0 | 0 | 0 | 0 |
| 1 | 0 | 1 | 0 | 0 | 0 | 0 | 0 |

### 7.2.1.3 Rotation gate type-SIII

In [17], [20], the following rotation gate is employed, where the number of conditions is as same as that of Table 7 and presented in Table 8. However, unlike the previous variant (type-II, subsection 6.2.2), the magnitude is a user-defined value, $\rho$ .

**Table 8: Lookup table for the rotation table type-SIII**

| $X_i$ | $b_i$ | F(X) <= F(b) | $\Delta\ \theta_i$ | s($\alpha_i$,$\beta_i$)) | | | |
|---|---|---|---|---|---|---|---|
| | | | | $\alpha_i\beta_i$> 0 | $\alpha_i\beta_i$< 0 | $\alpha_i$=0 | $\beta_i$=0 |
| 0 | 0 | 0 | 0 | 0 | 0 | 0 | 0 |
| 0 | 0 | 1 | 0 | 0 | 0 | 0 | 0 |
| 0 | 1 | 0 | $\rho$ | 1 | -1 | 0 | +/- 1 |
| 0 | 1 | 1 | $\rho$ | -1 | 1 | +/- 1 | 0 |
| 1 | 0 | 0 | $\rho$ | -1 | 1 | +/- 1 | 0 |
| 1 | 0 | 1 | $\rho$ | 1 | -1 | 0 | +/- 1 |
| 1 | 0 | 0 | 0 | 0 | 0 | 0 | 0 |
| 1 | 0 | 1 | 0 | 0 | 0 | 0 | 0 |

### 7.2.1.4 Rotation gate type-SIV

In [21], the authors have incorporated lookup table, shown in Table 9, to decide the magnitude of the rotation angle. Unlike the previous two variants, the sign is not determined through a lookup table. However, they ensure that there is a small rotation angle for each condition.

**Table 9: Lookup table for the rotation table type-SIV**

| $X_i$ | $b_i$ | F(X) > F(b) | $\Delta\ \theta_i$ |
|---|---|---|---|
| 0 | 0 | 0 | 0.001 $\pi$ |
| 0 | 0 | 1 | 0.001 $\pi$ |
| 0 | 1 | 0 | 0.08 $\pi$ |
| 0 | 1 | 1 | 0.001 $\pi$ |
| 1 | 0 | 0 | 0.08 $\pi$ |
| 1 | 0 | 1 | 0.001 $\pi$ |
| 1 | 0 | 0 | 0.001 $\pi$ |
| 1 | 0 | 1 | 0.001 $\pi$ |

Here, $X$ is the current solution and $b$ is the best solution

### 7.2.1.5 Rotation gate type-SV [31][33]

This type of rotation gate is considered to be a step-update based variant of that discussed in previous subsections, namely, type-SIII and type-SIV and presented in Table 10. Here, instead of a predefined value, the magnitude of the rotation is a user-defined parameter.

**Table 10: Lookup table for the rotation table type-SV**

| $X_i$ | $b_i$ | F(X) > F(b) | $\Delta\ \theta_i\ \rho$ | s($\alpha_i$,$\beta_i$)) | | | |
|---|---|---|---|---|---|---|---|
| | | | | $\alpha_i\beta_i$> 0 | $\alpha_i\beta_i$> 0 | $\alpha_i\beta_i$> 0 | $\alpha_i\beta_i$> 0 |
| 0 | 0 | 0 | 0 | 0 | 0 | 0 | 0 |
| 0 | 0 | 1 | 0 | 0 | 0 | 0 | 0 |
| 0 | 1 | 0 | 0 | 0 | 0 | 0 | 0 |
| 0 | 1 | 1 | $\rho$ | 1 | -1 | 0 | +/- 1 |
| 1 | 0 | 0 | 0 | 0 | 0 | 0 | 0 |
| 1 | 0 | 1 | $\rho$ | -1 | 1 | +/- 1 | 0 |
| 1 | 0 | 0 | 0 | 0 | 0 | 0 | 0 |
| 1 | 0 | 1 | 0 | 0 | 0 | 0 | 0 |

### 7.2.1.6 Rotation gate type-SVI

In [32], rotation gate type-SVI is presented where the magnitude of the rotation angle is based on the global best and the current solution. If the binary-encoded solution of the solution is true then the sign of the rotation is clock-wise, otherwise, the sign of the rotation is opposite (anti-clockwise). The corresponding equation is presented in Eq. 7.

$$\text{Xgb(y)\&X}_\text{I}\text{(y)} == 1)\ \ \theta\ = \theta \qquad (7)$$
$$\text{else}\quad \theta = -\ \theta$$

where Xgb is the global best and $X_I$ is the iteration best solution

### 7.2.1.7 Rotation gate type-SVII

In [51], [54], the authors have the employed lookup table (refer to Table 11). This is another form of lookup table, where the majority of the conditions have zero rotation angle. However, there were two conditions where the angle of the displacement is 0.01 radians and the sign of the displacement is based on the binary encoded values of the current and best solution.

**Table 11: Lookup table for the rotation table type-SVII**

| $X_i$ | $b_i$ | F(X) > F(b) | $\Delta\ \theta_i$ |
|---|---|---|---|
| | | | |
| 0 | 0 | 0 | 0 |
| 0 | 0 | 1 | 0.01 $\pi$ |
| 0 | 1 | 0 | -0.01 $\pi$ |
| 0 | 1 | 1 | 0 |
| 1 | 0 | 0 | 0 |
| 1 | 0 | 1 | 0 |
| 1 | 0 | 0 | 0 |
| 1 | 0 | 1 | 0 |

### 7.2.1.8 Rotation gate type-SVIII [31][33]

This type of rotation gate is considered to be a step-update based variant (see Table 12) of that discussed in previous subsections, namely, type-SIII and type-SIV. Here, instead of a predefined value, the magnitude of the rotation is a user-defined parameter.

**Table 12: Lookup table for the rotation table type-SVII**

| $X_i$ | $b_i$ | F(X) >= F(b) | $\Delta\,\theta_i$ | s($\alpha_i$,$\beta_i$)) | | | |
|---|---|---|---|---|---|---|---|
| | | | | $\alpha_i\beta_i$> 0 | $\alpha_i\beta_i$> 0 | $\alpha_i\beta_i$> 0 | $\alpha_i\beta_i$> 0 |
| 0 | 0 | 0 | 0.02 $\pi$ | -1 | 1 | 0 | 0 |
| 0 | 0 | 1 | 0 | 0 | 0 | 0 | 0 |
| 0 | 1 | 0 | 0.03 $\pi$ | -1 | 1 | +/- 1 | 0 |
| 0 | 1 | 1 | 0 | -1 | 1 | +/- 1 | 0 |
| 1 | 0 | 0 | 0.01 $\pi$ | 0 | 0 | 0 | 0 |
| 1 | 0 | 1 | 0 | 1 | -1 | 0 | +/- 1 |
| 1 | 0 | 0 | 0.05 $\pi$ | 1 | -1 | 0 | +/- 1 |
| 1 | 0 | 1 | 0 | 1 | -1 | 0 | +/- 1 |

### 7.2.2 Dynamic rotation gates

### 7.2.2.1 Rotation gate type-DI

In [22], the authors incorporated a user-defined angle which is predetermined, and all the Qubits evolved as per this angle.

### 7.2.2.2 Rotation gate type-DII

In [24], the authors developed an adaptive rotation angle, which is determined by the employed chaotic maps. The authors have employed seven chaotic maps, and the magnitude of the angle is based on the value generated through the underlying chaotic map as given in Eq. 8.

$$\Delta\,\theta(t)\ = \pi\,.\,\lambda\,(t-1) \tag{8}$$

For single-dimensional chaotic maps, $\lambda\,(t-1)$ is the value obtained from the chaotic map;
For two-dimensional chaotic maps, $\lambda\,(\,-1)$ is the absolute value of the output of the map in the x-axis over the maximum value of both generated chaotic series.

Here, $\lambda\,(t-1) = \frac{abs(\lambda_x(t))}{\max(\lambda_{x,y}\ (t)}$

### 7.2.2.3 Rotation gate type-DIII

In [25], the authors developed an adaptive rotation angle by considering the maximum and minimum values of the angle to be rotated and the elapse of the generation information. The adaptive variant is presented in Eq. 9.

$$\theta_i\ =\ \Delta\,\theta_i \tag{9}$$

where $\Delta\,\theta_i\ = \theta_{max}\ - \frac{(\theta_{max} - \theta_{min})x\ g}{G_{max}}$ ; g and $G_{max}$ denote the current and maximum number of generations, $\theta_{max}$ = 0.05 $\pi$ and $\theta_{min}$ = 0.0025 $\pi$ were the maximum and minimum rotation angles.

### 7.2.2.4 Rotation gate type-DIV [28]

This variant could be considered as a variant of that discussed in the previous subsection. However, while determining the magnitude, the fitness values were considered in this variant.

$$\Delta\theta_j = \theta \min \left( \theta\text{min_max} \frac{|F(x_j) - F(b)|}{\max\left(F(x_j), F(b)\right)} \right)_{\text{min}} \tag{10}$$

where $\theta_{min}$ and $\theta_{max}$ were the minimum and maximum values set by the user for the rotation angle, and F is the objective function score.

### 7.2.2.5 Rotation gate type-DV

This is another adaptive version of determining the rotation angle. Here, the rotation angle is determined through the Eq. 21 [49]. However, to determine the sign of the rotation angle, the authors have relied on the lookup table (see Table 13).

**Table 13: Lookup table for the rotation table type-XI**

| $X_i$ | $b_i$ | F(X) >= F(b) | s($\alpha_i$,$\beta_i$)) | | | |
|---|---|---|---|---|---|---|
| | | | $\alpha_i\beta_i$> 0 | $\alpha_i\beta_i$> 0 | $\alpha_i\beta_i$> 0 | $\alpha_i\beta_i$> 0 |
| 0 | 0 | 1 | -1 | +/-1 | 0 | +/- 1 |
| 0 | 0 | 0 | | | | |
| 0 | 1 | 1 | | | | |
| 0 | 1 | 0 | | | | |
| 1 | 0 | 0 | +/- 1 | -1 | +/- 1 | 0 |
| 1 | 0 | 1 | | | | |
| 1 | 0 | 0 | | | | |
| 1 | 0 | 1 | | | | |

## 7.3 Summary of the Survey

Overall, the following is the summary of the survey:

a) Majority of the studies have employed Quantum principles with an aim to enhance the convergence and diversity which is also achieved in obtaining better optimal solutions.

b) One of the most popular gates employed is rotation gate. However, the researchers did not follow any unified rule while employing rotation gate. The angle of rotation leads the evolution of the algorithm. A few of the works employs static rotation angle (see Section 7.2.1). However, making the angle adjust adaptively is highly essential. These kind of adaptive rotation angle is also noticed in a few studies (see Section 7.2.2).

c) Even though researchers have employed Q-bit notation. The probabilistic nature is not completely explored since only one gate (i.e., rotation gate) limiting the Quantum inference applicability.

d) The objective function also need to be chosen utmost careful. Majority of the researchers have employed the combined metric of cardinality of the feature subset into consideration and the accuracy.

## 8. Open Research Problems / Research Gaps

After conducting the survey, we identified the following potential open research problems:

- ***Employing other EAs:*** The Quantum-inspired versions of the other competing EAs based on novel metaphors could be parallelized and employed for FSS. It is a promising werea of research but the results could be subject to the No-Free-Lunch-Theorem [81].
- ***Hybridizing EAs of the above quantum operators:*** Developing hybrid QIEAs to balance the algorithms' exploration and exploitation capabilities of the algorithm is an interesting research direction to pursue.
- ***Employing Other Gates***: Each quantum gate uniquely performs the superposition of the data. Employing other quantum gates instead of rotation gates is another important future direction.
- ***Employing Quantum Entanglement Principle:*** Quantum entanglement [82] offers the real power of quantum computing to the EAs, if it is implemented during the exploration stage of the EAs even though it is memory-intensive.
- ***Hybridizing Reinforcement Learning and QIEAs:*** Reinforcement learning (RL) [83] can play a significant role in enhancing both the exploration and exploitation capabilities of QIEAs by virtue of its sequential decision making feature. Hence, hybridizing RL with QIEAs in either a loosely coupled or tightly coupled manner is a promising research direction. Deep RL can also be employed here but the associated computational costs in training the deep neural networks inherently present in the could be a show-stopper.
- ***Online Quantum FSS algorithms (both data and feature streams):*** Extending the works of traditional QIEAs under static settings to streaming data [73-74] is a very interesting future work.
- ***Handling unbalanced datasets***: Owing to the true capabilities of QIEAs, they could handle unbalanced datasets more efficiently than traditional EAs.
- ***Developing EAs in Quantum computers is a challenging task***: Extending the QIEAs, which were developed in classical computers to implement in Quantum computers such as Qiskit[1] and D-Wave system[2] etc., truly unleashes the power of exploration and exploitation.

## 9. Conclusions

This paper comprehensively surveys the existing quantum-inspired evolutionary algorithms (QIEAs) as applied to the Feature Subset Selection (FSS) problem. We discussed several time-tested and popular algorithms, highlighting their unique approaches. We also identified gaps in current research and suggested potential directions for future studies, , emphasizing the need for more robust and efficient solutions capable of handling high-dimensional data and dynamic environments.

Additionally, we present open research problems involving QIEAs for FSS. This review contains various QIEA aspects, providing a more detailed taxonomy to offer a thorough understanding and a solid foundation for future research in this werea.

[1] Qiskit : https://github.com/Qiskit/qiskit

[2] D-Wave https://docs.dwavesys.com/docs/latest/index.html